\documentclass[acmtog]{acmart}
\acmSubmissionID{411}

\usepackage{booktabs} 
\usepackage{multirow}
\usepackage{float}

\usepackage[ruled]{algorithm2e} 

\SetAlFnt{\small}
\SetAlCapFnt{\small}
\SetAlCapNameFnt{\small}
\SetAlCapHSkip{0pt}

\acmJournal{TOG}
\acmVolume{39}
\acmNumber{6}
\acmArticle{235}
\acmYear{2020}
\acmMonth{12}

\setcopyright{rightsretained}

\acmDOI{10.1145/3414685.3417877}

\definecolor{myblue}{rgb}{0.188, 0.188, 0.858}
\definecolor{darkorange}{rgb}{1.0, 0.55, 0.0}

\newcommand{\ignore}[1]{{}}

\def\bfE{{\bf E}}

\newcommand{\physcap}{\textit{PhysCap}}

\begin{document}
\title{PhysCap: Physically Plausible Monocular 3D Motion Capture in Real Time} 
\thanks{This work was funded by the ERC Consolidator Grant 4DRepLy (770784)}
\author{Soshi Shimada}
\affiliation{%
 \institution{Max Planck Institute for Informatics, Saarland Informatics Campus }
 }
 \email{sshimada@mpi-inf.mpg.de}
 \author{Vladislav Golyanik}
\affiliation{%
 \institution{Max Planck Institute for Informatics, Saarland Informatics Campus}
 }
 \email{golyanik@mpi-inf.mpg.de}
 \author{Weipeng Xu}
\affiliation{%
 \institution{Facebook Reality Labs}
 }
 \email{xuweipeng@fb.com}
 \author{Christian Theobalt}
\affiliation{%
 \institution{Max Planck Institute for Informatics, Saarland Informatics Campus }
 } 
 \email{theobalt@mpi-inf.mpg.de}
\begin{abstract} 
Marker-less 3D human motion capture 
from a single colour camera has seen significant  progress. 
However, it is a very challenging and severely ill-posed problem. 
In consequence, even the most accurate state-of-the-art approaches have significant limitations. 
Purely kinematic formulations on the basis of individual joints or skeletons, and the frequent frame-wise reconstruction in state-of-the-art methods greatly limit 3D accuracy and temporal stability compared to multi-view or marker-based motion capture. 
Further, captured 3D poses are often physically incorrect and biomechanically implausible, or exhibit implausible environment interactions (floor penetration, foot skating, unnatural body leaning and strong shifting in depth), which is problematic for any use case in computer graphics. 
We, therefore, present \physcap{}, the first algorithm for physically plausible, real-time and marker-less human 3D motion capture with a single colour camera at $25$~fps. 
Our algorithm first captures 3D human poses purely kinematically. 
To this end, a CNN infers 2D and 3D joint positions, and subsequently, an inverse kinematics step finds space-time coherent joint angles and global 3D pose. 
Next, these kinematic reconstructions are used as constraints in a real-time physics-based pose optimiser that accounts for environment constraints (\textit{e.g.,} collision handling and floor placement), gravity, and biophysical plausibility of human postures. 
Our approach employs a combination of ground reaction force and residual force for plausible root control, and uses a trained neural network to detect foot contact events in images. 
Our method captures physically plausible and temporally stable global 3D human motion, without physically implausible postures, floor penetrations or foot skating, from video in real time and in general scenes.
\physcap{} achieves state-of-the-art accuracy on established pose benchmarks, and we propose new metrics to 
demonstrate the improved physical plausibility and temporal stability. 
\end{abstract}

\begin{teaserfigure}
\centering
\includegraphics[width=\linewidth]{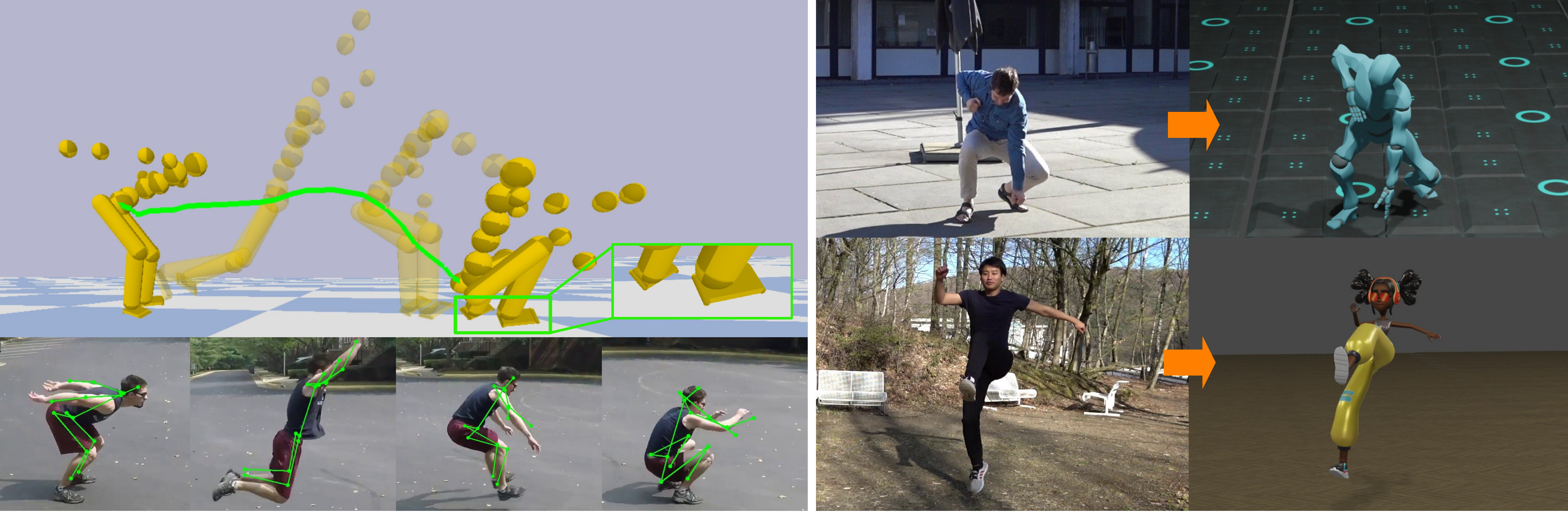}
\caption{ 
\physcap{} captures global 3D human motion in a physically plausible way from monocular videos in real time, automatically and without the use of markers.  
(Left:) Video of a \textit{standing long jump} \cite{Peng2018} and our 3D reconstructions. Thanks to its formulation on the basis of physics-based dynamics, our algorithm recovers challenging 3D human motion observed in 2D while significantly mitigating artefacts such as foot sliding, foot-floor penetration, unnatural body leaning and jitter along the depth channel that troubled earlier monocular pose estimation methods. 
(Right:) Since the output of \physcap{} is environment-aware and the returned root position is global, it is directly suitable for virtual character animation, without any further post-processing.
The 3D characters are taken from \cite{Mixamo}. 
See our supplementary video for further results and visualisations. 
} 
\label{fig:teaser} 
\end{teaserfigure} 
%
%


%
%


\maketitle

\section{Introduction} 
3D human pose estimation from monocular RGB images is a very active area of research. 
Progress is fueled by many applications with an increasing need for reliable, real time and simple-to-use pose estimation. 
Here, applications in character animation, VR and AR, telepresence, or human-computer interaction, are only a few examples of high importance for graphics.

Monocular and markerless 3D capture of the human skeleton is a highly challenging and severely underconstrained problem \cite{VNect_SIGGRAPH2017, JMartinezICCV_2017,  Pavlakos2018OrdinalDS, Kovalenko2019, WandtRosenhahn2019}. 
Even the best state-of-the-art algorithms, therefore, exhibit notable limitations. 
Most methods capture pose kinematically using individually predicted joints but do not produce smooth joint angles of a coherent kinematic skeleton. 
Many approaches perform per-frame pose estimates with notable temporal jitter, and reconstructions are often in root-relative but not global 3D space. 
Even if a global pose is predicted, depth prediction from the camera is often unstable. 
Also, interaction with the environment is usually entirely ignored, which leads to poses with severe collision violations, \textit{e.g.,}  floor penetration  or the implausible foot sliding and incorrect foot placement. 
Established kinematic formulations also do not explicitly consider biomechanical plausibility of reconstructed poses, yielding reconstructed poses with improper balance, inaccurate body leaning, or temporal instability. 

We note that all these artefacts are particularly problematic in the aforementioned computer graphics applications, in which temporally stable and visually plausible motion control of characters from all virtual viewpoints, in global 3D, and with respect to the physical environment, are critical. 
Further on, we note that established metrics in widely-used 3D pose estimation benchmarks \cite{ionescu2013human3, mono-3dhp2017}, such as mean per joint position error (MPJPE) or 3D percentage of correct keypoints (3D-PCK), which are often even evaluated after a 3D rescaling or Procrustes alignment, do not adequately measure these artefacts. 
In fact, we show (see Sec.~\ref{sec:method}, and supplemental video) that even some top-performing methods on these benchmarks produce results with substantial temporal noise and unstable depth prediction, with frequent violation of environment constraints, and with frequent disregard of physical and anatomical pose plausibility. 
In consequence, there is still a notable gap between monocular 3D pose human estimation approaches and the gold standard accuracy and motion quality of suit-based or marker-based motion capture systems, which are unfortunately expensive, complex to use and not suited for many of the aforementioned applications requiring in-the-wild capture. 

We, therefore, present \physcap{} -- a new approach for easy-to-use monocular global 3D human motion capture that significantly narrows this gap and substantially reduces the aforementioned artefacts, see Fig.~\ref{fig:teaser} for an overview. 
\physcap{} is, to our knowledge, the first method that jointly possesses all the following properties: it is fully-automatic, markerless, works in general scenes, runs in real time, captures a space-time coherent skeleton pose and global 3D pose sequence of state-of-the-art temporal stability and smoothness. 
It exhibits state-of-the-art posture and position accuracy, and captures physically and anatomically plausible poses that correctly adhere to physics and environment constraints. 
To this end, we rethink and bring together in new way ideas from kinematics-based monocular pose estimation and physics-based human character animation. 

The \emph{first stage} of our algorithm is similar to \cite{VNect_SIGGRAPH2017} and estimates 3D body poses in a purely kinematic, physics-agnostic way. 
A convolutional neural network (CNN) infers combined 2D and 3D joint positions from an input video, which are then refined in a space-time inverse kinematics to yield the first estimate of skeletal joint angles and global 3D poses. 
In the \emph{second stage}, the foot contact and the motion states are predicted for every frame. 
Therefore, we employ a new CNN that detects heel and forefoot placement on the ground from estimated 2D keypoints in images, and classifies the observed poses into stationary or non-stationary. 
In the \emph{third stage}, the final physically plausible 3D skeletal joint angle and pose sequence is computed in real time. 
This stage regularises human motion with a torque-controlled physics-based character represented by a kinematic chain with a floating base.
To this end, the optimal control forces for each degree of freedom (DoF) of the kinematic chain are computed, such that the kinematic pose estimates from the first stage -- in both 2D and 3D -- are reproduced as closely as possible. 
The optimisation ensures that physics constraints like gravity, collisions, foot placement, as well as physical pose plausibility (\textit{e.g.,} balancing), are fulfilled. 
To summarise, our \textbf{contributions} in this article are: 
\begin{itemize} 
    \item The first, to the best of our knowledge, marker-less monocular 3D human motion capture approach on the basis of an explicit physics-based dynamics model which runs in real time and captures global, physically plausible skeletal motion (Sec.~\ref{sec:method}). 
    \item A CNN to detect foot contact and motion states from images (Sec.~\ref{sec:method_stage_2}). 
    \item A new pose optimisation framework with a human 
    parametri- sed by a torque-controlled simulated character with a floating base and PD joint controllers; it reproduces kinematically captured 2D/3D poses and simultaneously accounts for physics constraints like ground reaction forces, foot contact states and collision response (Sec.~\ref{sec:method_stage_3}). 
    \item Quantitative metrics to assess frame-to-frame jitter and floor penetration in captured motions (Sec.~\ref{ssec:evaluation_methodology}). 
    \item Physically-justified results with significantly fewer artefacts, such as frame-to-frame jitter, incorrect leaning, foot sliding and floor penetration than related methods (confirmed by a user study and metrics), as well as state-of-the-art 2D and 3D accuracy and temporal stability (Sec.~\ref{sec:results}). 
\end{itemize} 

We demonstrate the benefits of our approach through experimental evaluation on several datasets  (including newly recorded videos) against multiple state-of-the-art methods for monocular 3D human motion capture and pose estimation. 
\section{Related Work}\label{sec:related_works} 
Our method mainly relates to two different categories of approaches --   
(markerless) 3D human motion capture from colour imagery, and physics-based character animation. 
In the following, we review related types of methods, focusing on the most closely related works. 

\paragraph{Multi-View Methods for 3D Human Motion Capture from RGB} 
Reconstructing humans from multi-view images is well studied. 
Multi-view motion capture methods track the articulated skeletal motion, usually by fitting an articulated template to imagery~\cite{StollHGST11, Bro10b, bo_ijcv08, Brox10, outdoorsHMC, Marconi, outdoormocapTVCG,20204DAssociation}. 

Other methods, sometimes termed performance capture methods, additionally capture the non-rigid surface deformation, \textit{e.g.,} of clothing ~\cite{starck2007surface,waschbusch2005scalable,vlasic2009dynamic,cagniart2010free}. 
They usually fit some form of a template model to multi-view imagery~\cite{de2008performance,bradley2008markerless,Brualla2018} that often also has an underlying kinematic skeleton  ~\cite{gall2009motion,vlasic2008articulated,liu2011markerless,wu2012full}. 
Multi-view methods have demonstrated compelling results and some enable free-viewpoint video. 
However, they require expensive multi-camera setups and often controlled studio environments. 

\paragraph{Monocular 3D Human Motion Capture and Pose Estimation from RGB} 
Marker-less 3D human pose estimation (reconstruction of 3D joint positions only) and motion capture (reconstruction of global 3D body motion and joint angles of a coherent skeleton) from a single colour or greyscale image are highly ill-posed problems. 
The state of the art on monocular 3D human pose estimation has greatly progressed in recent years, mostly fueled by the power of trained CNNs~\cite{inthewild3d_2019,mono-3dhp2017}. 
Some methods estimate 3D pose by combining 2D keypoints prediction with body depth regression~\cite{Zhou_2017_ICCV,Newell2016StackedHN,Yang_3dposeCVPR2018,Dabral:ECCV:2018} or with regression of 3D joint location probabilities~\cite{pavlakos2017volumetric,VNect_SIGGRAPH2017} in a trained CNN. Lifting methods predict joint depths from detected 2D keypoints~\cite{tome2017lifting,chen_2017_3d,JMartinezICCV_2017,Pavlakos2018OrdinalDS}.
Other CNNs regress 3D joint locations directly~\cite{Tekin2016StructuredPO,mono-3dhp2017,rhodin2018unsupervised}.
Another category of methods combines CNN-based keypoint detection with constraints from a parametric body model, \textit{e.g.}, by using reprojection losses during training~\cite{Bogo:ECCV:2016,brau_2016_3d,inthewild3d_2019}.
Some works approach monocular multi-person 3D pose estimation~\cite{RogezWS18} and motion capture~\cite{XNect_SIGGRAPH2020}, or estimate non-rigidly deforming human surface geometry from monocular video on top of skeletal motion~\cite{Habermann:2019,deepcap,EventCap2020}.
In addition to greyscale images, \cite{EventCap2020} use an asynchronous event stream from an event camera as input. 
Both these latter directions are complementary but orthogonal to our work. 

The majority of methods in this domain estimates 3D pose as a  root-relative 3D position of the body joints  \cite{JMartinezICCV_2017, 
Moreno-Noguer_2017, Pavlakos2018OrdinalDS, WandtRosenhahn2019,  Kovalenko2019}. 
This is problematic for applications in graphics, as temporal jitter, varying bone lengths and the often not recovered global 3D pose make animating virtual characters hard. 
Other monocular methods are trained to estimate parameters or joint angles of a skeleton~\cite{zhou2016deep} or parametric model \cite{hmrKanazawa17}. 
\cite{VNect_SIGGRAPH2017, XNect_SIGGRAPH2020} employ inverse kinematics on top of CNN-based 2D/3D inference to obtain joint angles of a coherent skeleton in global 3D and in real-time. 

Results of all aforementioned methods frequently violate laws of physics, and exhibit foot-floor penetrations, foot sliding, and unbalanced or implausible poses floating in the air, as well as notable jitter. 
Some methods try to reduce jitter 
by exploiting temporal information \cite{humanMotionKanazawa19, kocabas2019vibe}, \textit{e.g.,} by estimating smooth multi-frame scene trajectories~\cite{Peng2018}.
\cite{Zou2020} try to reduce foot sliding by ground contact constraints. 
\cite{Zanfir2018} jointly reason about ground planes and volumetric occupancy for multi-person pose estimation. 
\cite{Monszpart2019} jointly infer coarse scene layout and human pose from monocular interaction video, and~\cite{PROX:2019} use a pre-scanned 3D model of scene geometry to constrain kinematic pose optimisation. 
To overcome the aforementioned limitations, no prior work formulates monocular motion capture on the basis of an explicit physics-based dynamics model and in real-time, as we do.
\paragraph{Physics-Based Character Animation.} 
Character animation on the basis of physics-based controllers has been investigated for many years \cite{Barzel1996, Sharon2005Walking, Wrotek2006}, and remains an active area of research,  \cite{levine2012physically, zheng2013human,  andrews2016real, Bergamin2019}. 
\cite{levine2012physically} employ a quasi-physical simulation that approximates a reference motion trajectory in real-time. They can follow non-physical reference motion by applying a direct actuation at the root. 
By using proportional derivative (PD) controllers and computing optimal torques and contact forces, \cite{zheng2013human} make a character follow a reference motion captured while keeping balance. 
\cite{Liu2010Samcon} proposed a probabilistic algorithm for physics-based character animation. 
Due to the stochastic property and inherent randomness, their results evince variations, but the method requires multiple minutes of runtime per sequence. 
Andrews \textit{et al.}~\shortcite{andrews2016real} employ rigid dynamics to drive a virtual character from a combination of marker-based motion capture and body-mounted sensors. 
This animation setting is related to motion transfer onto robots. 
\cite{Nakaoka2007} transferred human motion captured by a multi-camera marker-based system onto a robot, with an emphasis on leg motion.
\cite{zhang2014leveraging} leverage depth cameras and wearable  pressure sensors and apply physics-based motion optimisation. 
We take inspiration from these works for our setting, where we have to capture in a physically correct way and in real time global 3D human motion from images, using intermediate pose reconstruction results that exhibit notable artefacts and violations of physics laws. 
\physcap{}, therefore, combines an initial kinematics-based pose reconstruction with PD controller based physical pose optimisation. 

Several recent methods apply deep reinforcement learning to virtual character animation control~\cite{Peng2018, Bergamin2019, Lee2019}. 
Peng \textit{et al.}~\shortcite{Peng2018} propose a reinforcement learning approach for transferring dynamic human performances observed in monocular videos. 
They first estimate smooth motion trajectories with recent monocular human pose  estimation techniques, and then train an imitating control policy for a virtual  character. 
\cite{Bergamin2019} train a controller for a virtual character from several minutes of motion capture data which covers the expected variety of motions and poses. 
Once trained, the virtual character can follow directional commands of the user in real time, while being robust to collisional obstacles. 
Other work \cite{Lee2019} combines a muscle actuation model with deep reinforcement learning. 
\cite{Jiang2019} express an animation objective in muscle actuation space. 
The work on learning animation controllers for specific motion classes is inspirational but different from real-time physics-based motion capture of general motion.
\paragraph{Physically Plausible Monocular 3D Human Motion Capture.} 
Only a few works on monocular 3D human motion capture using explicit physics-based constraints  exist~\cite{wei2010videomocap,vondrak2012video,ZelWan2017,li2019motionforcesfromvideo}. 
\cite{wei2010videomocap} capture 3D human poses from uncalibrated monocular video using physics constraints. 
Their approach requires manual user input for each frame of a video. In contrast, our approach is automatic, runs in real time, and uses a different formulation for physics-based pose optimisation geared to our setting. %
\cite{vondrak2012video} capture bipedal controllers from a video. Their controllers are robust to perturbations and generalise well for a variety of motions. However, unlike our \physcap{}, the generated motion often looks unnatural and their method does not run in real time. 
\cite{ZelWan2017} capture poses and internal body forces from images only for certain classes of motion (\textit{e.g.,} lifting and walking) by using a data-driven approach, but not an explicit forward dynamics approach handling a wide range of motions, like ours. 

Our \physcap{} bears most similarities with the rigid body dynamics based monocular human pose estimation by Li~\textit{et al.}~\shortcite{li2019motionforcesfromvideo}. 
Li~\textit{et al.} estimate 3D poses, contact states and forces from input videos with physics-based constraints. 
However, their method and our approach are substantially different. 
While Li~\textit{et al.} focus on object-person interactions, we target a variety of general motions, including complex acrobatic motions such as backflipping without objects. 
Their method does not run in real time and requires manual annotations on images to train the contact state estimation networks. 
In contrast, we leverage the PD controller based inverse dynamics  tracking, which results in physically plausible, smooth and natural skeletal pose and root motion capture in real time. 
Moreover, our contact state estimation network relies on annotations generated in a semi-automatic way. 
This enables our architecture to be trained on large datasets, which results in the improved generalisability. 
No previous method of the reviewed category ``physically plausible monocular 3D human motion capture''  combines \textit{the ability of our algorithm to capture global 3D human pose of similar quality and physical plausibility in real time}. 

\section{Body Model and Preliminaries}\label{sec:model} 
\begin{figure}[t!] 
\centering 
\includegraphics[width=\linewidth]{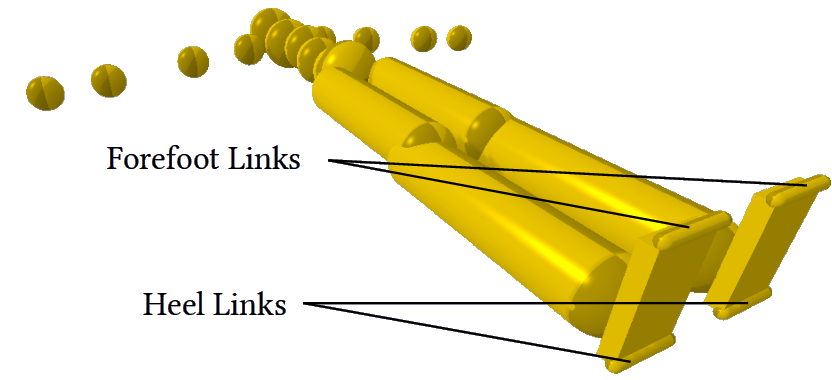} 
\caption{Our virtual character used in stage III. 
The forefoot and heel links are involved in the mesh  collision checks with the floor plane in the physics engine \cite{coumans2016pybullet}. 
} 
\label{fig:footlinks} 
\end{figure} 
The input to \physcap{} is a 2D image sequence $\mathbf{I}_{t}$, $t \in \{1,  \hdots, T\}$, where $T$ is the total number of frames and $t$ is the frame index. 
We assume a perspective camera model and calibrate the camera and floor location before tracking starts. 
Our approach outputs a physically plausible real-time 3D motion capture result $\mathbf{q}^{t}_{phys} \in \mathbb{R}^{m}$ (where  $m$ is the number of degrees of freedom) that adheres to the image observation, as well as physics-based posture and environment constraints.
For our human model, $m = 43$. Joint angles are parametrised by Euler angles. The mass distribution of our character is computed following \cite{Liu2010Samcon}.
Our character model has a skeleton composed of 37 joints and \textit{links}. A link defines the volumetric extent of a body part via a collision proxy. 
The forefoot and heel links, centred at the respective joints of our character (see Fig.~\ref{fig:footlinks}), are used to detect foot-floor collisions during physics-based pose optimisation. 

Throughout our algorithm, we represent 
the pose of our character by a combined vector $\mathbf{q} \in \mathbb{R}^{m}$~\cite{featherstone2014rigid}.
The first three entries of $\mathbf{q}$ contain the global 3D root position in Cartesian coordinates, the next three entries encode the orientation of the root, and the remaining entries are the joint angles. 
When solving for the physics-based motion capture result, 
the motion of the physics-based character will be controlled by the vector of forces denoted by $\boldsymbol{\tau} \in \mathbb{R}^{m}$ interacting with gravity, Coriolis and centripetal forces $\mathbf{c}\in \mathbb{R}^{m}$. The root of our character is not fixed and can globally move in the environment, which is commonly called a floating-base system. Let the velocity and acceleration of $\mathbf{q}$ be $\dot{\mathbf{q}}\in  \mathbb{R}^{m}$ and $\ddot{\mathbf{q}}\in \mathbb{R}^{m}$, respectively. Using the finite-difference method, the relationship between $\mathbf{q}, \dot{\mathbf{q}},\ddot{\mathbf{q}}$ can be written as 
\begin{equation} \label{eq:findif} 
\begin{aligned} 
    &\dot{\mathbf{q}}^{i+1}=\dot{\mathbf{q}}^{i}+\phi\ddot{\mathbf{q}}^{i},\\ 
    &\mathbf{q}^{i+1}= \mathbf{q}^{i} +\phi\dot{\mathbf{q}}^{i+1}, 
\end{aligned} 
\end{equation} 
where $i$ represents the simulation step index and $\phi = 0.01$ is the simulation step size. 

For the motion to be physically plausible, $\ddot{\mathbf{q}}$ and the vector of forces $\boldsymbol{\tau}$ must satisfy the equation of motion \cite{featherstone2014rigid}: 
\begin{equation} \label{eq:eom}
   \mathbf{M}(\mathbf{q}) \ddot{\mathbf{q}} - \boldsymbol{\tau}   =   \mathbf{J}^{T} \mathbf{G}\mathbf{\lambda} - \mathbf{c}(\mathbf{q},\dot{\mathbf{q}}),
\end{equation} 
where $\mathbf{M}\in \mathbb{R}^{m \times m}$ is a joint space inertia matrix which is composed of the moment of inertia of the system. 
It is computed using the Composite Rigid Body algorithm \cite{featherstone2014rigid}. 
$\mathbf{J}\in \mathbb{R}^{6N_{c}\times m}$ is a contact Jacobi matrix which relates the external forces to joint coordinates, with $N_{c}$ denoting the number of links where the contact force is applied. $\mathbf{G}\in \mathcal{R}^{6N_{c} \times 3N_{c}}$ transforms contact forces $\mathbf{\lambda}\in \mathcal{R}^{3N_{c}}$ into the linear force and torque \cite{zheng2013human}. 
 
Usually, in a floating-base system, the first six entries of $\boldsymbol{\tau}$ which correspond to the root motion are set to $0$ for a humanoid character control. 
\begin{figure*}[t!]
\centering
\includegraphics[width=\textwidth]{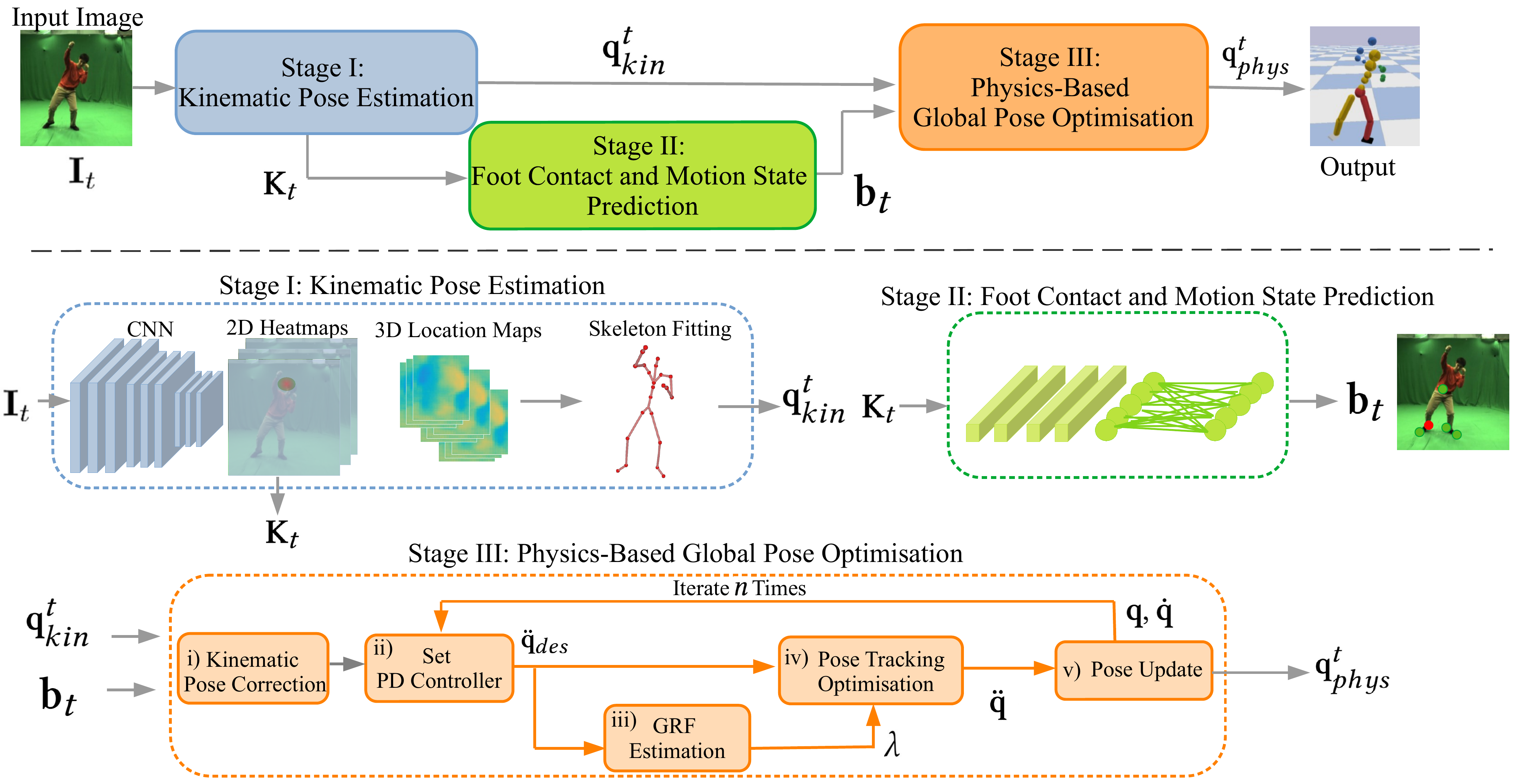}
\caption{\textbf{Overview of our pipeline.}  
In stage I, the 3D pose estimation network accepts RGB   image $\mathbf{I}_{t}$ as input and returns 2D joint keypoints $\mathbf{K}_{t}$ along with the global 3D pose  $\mathbf{q}^{t}_{kin}$, \textit{i.e.,} root translation,  orientation and joint angles of a kinematic skeleton. 
In stage II, $\mathbf{K}_{t}$ is fed to the contact and  motion state detection network. 
Stage II returns the contact states of heels and forefeet as well as a label  $\mathbf{b}_{t}$ that represents if the subject in  $\mathbf{I}_{t}$ is stationary or not. 
In stage III, $\mathbf{q}^{t}_{kin}$ and $\mathbf{b}_{t}$  are used to iteratively update the character pose  respecting physics laws. 
After the $n$ pose update iterations, we obtain the final  3D pose $\mathbf{q}^{t}_{phys}$. 
Note that the orange arrows in stage III represent the  steps that are repeated in the loop in every iteration. 
Kinematic pose correction is performed only once at the  beginning of stage III. 
}
\label{fig:pipeline}
\end{figure*} 
This reflects the fact that humans do not directly control root translation and orientation by muscles acting on the root, but indirectly by the other joints and muscles in the body. 
In our case, however, the kinematic pose $\mathbf{q}^{t}_{kin}$ which our final physically plausible result shall reproduce as much as possible (see Sec.~\ref{sec:method}), is estimated from a monocular image sequence (see stage I in Fig.~\ref{fig:pipeline}), which contains physically implausible artefacts. 
Solving for joint torque controls that blindly make the character follow,  would make the character quickly fall down. 
Hence, we keep the first six entries of $\boldsymbol{\tau}$ in our formulation and can thus directly control the root position and orientation with an additional external force. 
This enables the final character motion to keep up with the global root trajectory estimated in the first stage of \physcap{}, without falling down.

\section{Method}\label{sec:method} 

Our \physcap{} approach includes three stages, see Fig.~\ref{fig:pipeline} for an overview. 
The first stage performs \textit{kinematic pose estimation}. 
This encompasses 2D heatmap and 3D location map regression for each body joint with a CNN, followed by a model-based space-time pose optimisation step (Sec.~\ref{sec:method_stage_1}). 
This stage returns 3D skeleton pose in joint angles $\mathbf{q}^{t}_{kin} \in \mathbb{R}^{m}$ along with the 2D joint keypoints  $\mathbf{K}_{t} \in \mathbb{R}^{s\times 2}$ for every image; 
$s$ denotes the number of 2D joint keypoints. 
As explained earlier, this initial kinematic reconstruction $\mathbf{q}^{t}_{kin}$ is prone to physically implausible effects such as foot-floor penetration, foot skating, anatomically implausible body leaning and temporal jitter, especially notable along the depth dimension. 
The second stage performs \textit{foot contact and motion state detection,} which uses 2D joint detections $\mathbf{K}_{t}$ to classify the poses reconstructed so far into stationary and non-stationary -- this is stored in one binary flag. 
It also estimates binary foot-floor contact flags, \textit{i.e.,} for the toes and heels of both feet, resulting in four binary flags (Sec.~\ref{sec:method_stage_2}). 
This stage outputs the combined state vector $\mathbf{b}_{t} \in \mathbb{R}^{5}$.
The third and final stage of \physcap{} is the  \textit{physically plausible global 3D pose estimation} (Sec.~\ref{sec:method_stage_3}). 
It combines the estimates from the first two stages with physics-based constraints to yield a physically plausible real-time 3D motion capture result that adheres to physics-based posture and environment constraints $\mathbf{q}^{t}_{phys} \in \mathbb{R}^{m}$. 
In the following, we describe each of the stages in detail. 

\subsection{Stage I: Kinematic Pose  Estimation}\label{sec:method_stage_1} 
Our kinematic pose estimation stage follows the real-time VNect algorithm \cite{VNect_SIGGRAPH2017}, see Fig.~\ref{fig:pipeline}, stage I. 
We first predict heatmaps of 2D joints and root-relative location maps of joint positions in 3D with a specially tailored fully convolutional neural network using a ResNet \cite{he2016deep} core.  
The ground truth joint locations for training are taken from the MPII \cite{Andriluka2014} and LSP \cite{JohnsonEveringham2011} datasets in the 2D case, and MPI-INF-3DHP \cite{mono-3dhp2017} and Human3.6m \cite{ionescu2013human3} datasets in the 3D case. 

Next, the estimated 2D and 3D joint locations are temporally filtered and used as constraints in a kinematic skeleton fitting step that optimises the following energy function: 
\begin{equation}\label{eq:VNect} 
\begin{aligned}
 \bfE_{kin}(\mathbf{q}^{t}_{kin}) =  &\bfE_{\text{IK}}(\mathbf{q}^{t}_{kin}) + \bfE_{\text{proj.}}(\mathbf{q}^{t}_{kin}) \, + \\ &\bfE_{\text{smooth}}(\mathbf{q}^{t}_{kin}) + \bfE_{\text{depth}}(\mathbf{q}^{t}_{kin}).  
\end{aligned} 
\end{equation}
The energy function \eqref{eq:VNect} contains four terms (see~\cite{VNect_SIGGRAPH2017}), \textit{i.e.,} the 3D inverse kinematics term $\bfE_{\text{IK}}$, the projection term $\bfE_{\text{proj.}}$, the temporal stability term $\bfE_{\text{smooth}}$ and the depth uncertainty correction term $\bfE_{\text{depth}}$. 
$\bfE_{\text{IK}}$ is the data term which constrains the 3D pose to  be close to the 3D joint predictions from the CNN. 
$\bfE_{\text{proj.}}$ enforces the pose $\mathbf{q}^{t}_{kin}$ to  reproject it to the 2D keypoints (joints) detected by the CNN. 
Note that this reprojection constraint, together with calibrated camera and calibrated bone lengths, enables computation of the global 3D root (pelvis) position in the camera space. 
Temporal stability is further imposed by penalising the root's  acceleration and variations along the depth channel by  $\bfE_{\text{smooth}}$ and $\bfE_{\text{depth}}$, respectively. 
The energy \eqref{eq:VNect} is optimised by non-linear least squares (Levenberg-Marquardt algorithm \cite{Levenberg_44,Marquardt_1963}), and the obtained vector of joint angles and the root rotation and position  $\mathbf{q}^{t}_{kin}$ 
of a skeleton with fixed bone lengths are smoothed by an adaptive first-order low-pass filter \cite{Casiez2012}. 
Skeleton bone lengths of a human can be computed, up to a global scale, from averaged 3D joint detections of a few initial frames. 
Knowing the metric height of the human determines the scale factor to compute metrically correct global 3D poses. 
\begin{figure}[t!] 
\centering 
\includegraphics[width=\linewidth]{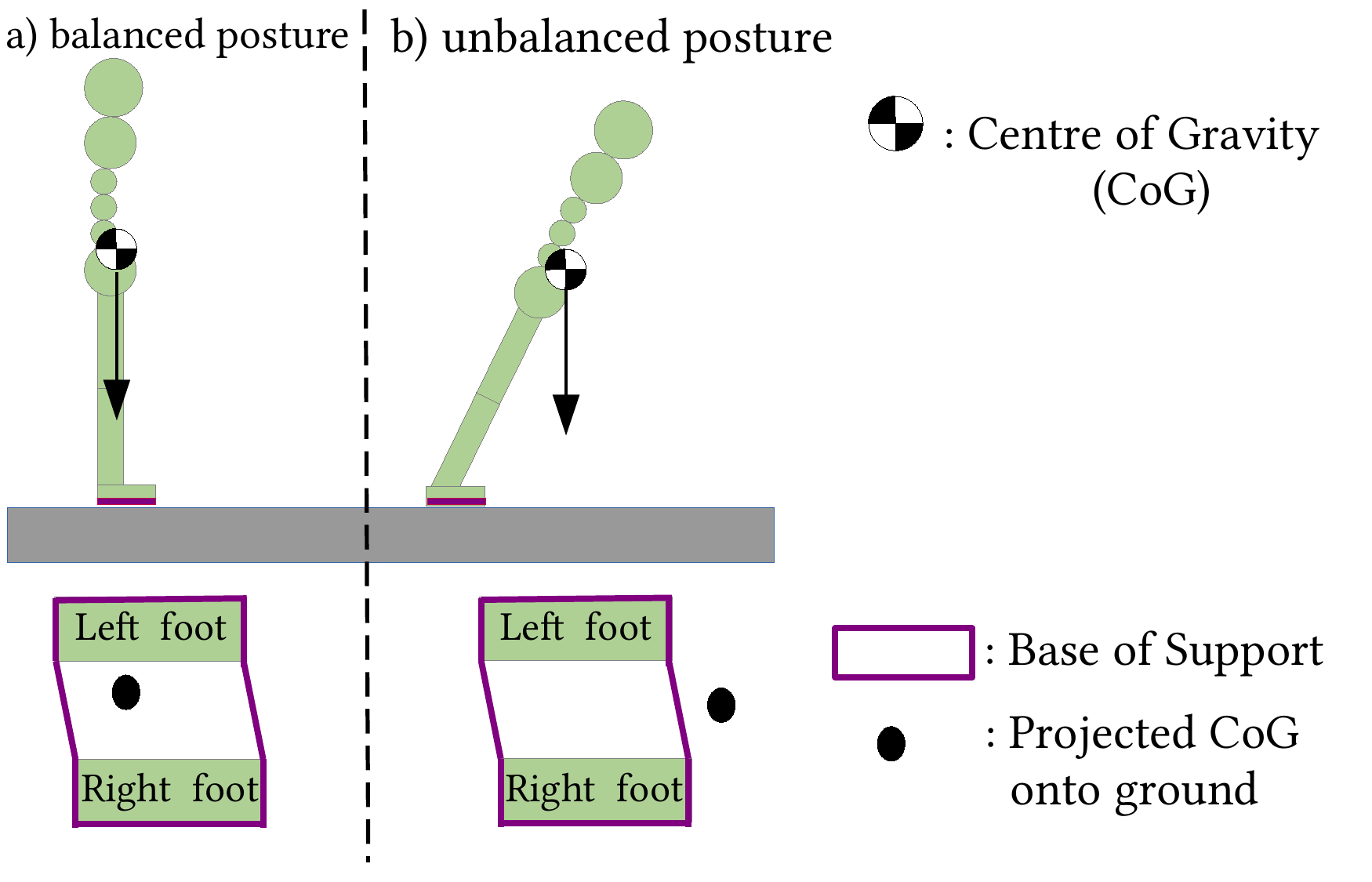} 
\caption{
(a) Balanced posture: the CoG of the body projects inside the base of support. (b) Unbalanced posture: the CoG does not project inside the base of support, which causes the human to start losing a balance.   } 
\label{fig:supportPolygon} 
\end{figure} 

The result of stage I is a temporally-consistent joint angle sequence but, as noted earlier, captured poses can exhibit artefacts and contradict physical plausibility (\textit{e.g.,} evince floor penetration, incorrect body leaning, temporal jitter, \textit{etc.}). 

\subsection{Stage II: Foot Contact and Motion State  Detection}\label{sec:method_stage_2} 
The ground reaction force (GRF) -- applied when the feet touch the ground -- enables humans to walk and control their posture.
The interplay of internal body forces and the ground reaction force controls human pose, which enables locomotion and body balancing by controlling the centre of gravity (CoG). 
To compute physically plausible poses accounting for the GRF in stage III, we thus need to know foot-floor contact states. 
Another important aspect of the physical plausibility of biped poses, in general, is balance. 
When a human is standing or in a stationary upright state, the CoG of her body projects inside a base of support (BoS). 
The BoS is an area on the ground bounded by the foot contact points, see Fig.~\ref{fig:supportPolygon} for a visualisation. 
When the CoG projects outside the BoS in a stationary pose, a human starts losing balance and will fall if no correcting motion or step is applied. 
Therefore, maintaining a static pose with an extensive leaning, as often observed in the results of monocular pose estimation, is not physically plausible (Fig~.\ref{fig:supportPolygon}-(b)). 
The aforementioned CoG projection criterion can be used to correct imbalanced stationary poses \cite{Faloutsos2001, Macchietto2009, Coros2010}. 
To perform such correction in stage III, we need to know if a pose is stationary or non-stationary  (whether it is a part of a locomotion/walking phase). 

Stage II, therefore, estimates foot-floor contact states of the feet in each frame and determines whether the pose of the subject in $\mathbf{I}_{t}$ is stationary or not. To predict both, \textit{i.e.,} foot contact and motion states, we use a neural network whose architecture extends Zou \textit{et al.} \shortcite{Zou2020} who only predict foot contacts. It is composed of temporal convolutional layers with one fully connected layer at the end. 
The network takes as input all 2D keypoints $\mathbf{K}_{t}$ from the last seven time steps (the temporal window size is set to seven), and returns for each image frame binary labels indicating whether the subject is in the stationary or non-stationary pose, as well as the contact state flags for the forefeet and heels of both feet encompassed in $\mathbf{b}_{t}$. 
The supervisory labels for training this network are automatically computed on a subset of the 3D motion sequences of the Human3.6M \cite{ionescu2013human3} and DeepCap \cite{deepcap} datasets using the following criteria: 
\begin{figure}[t!] 
\centering 
\includegraphics[width=\linewidth]{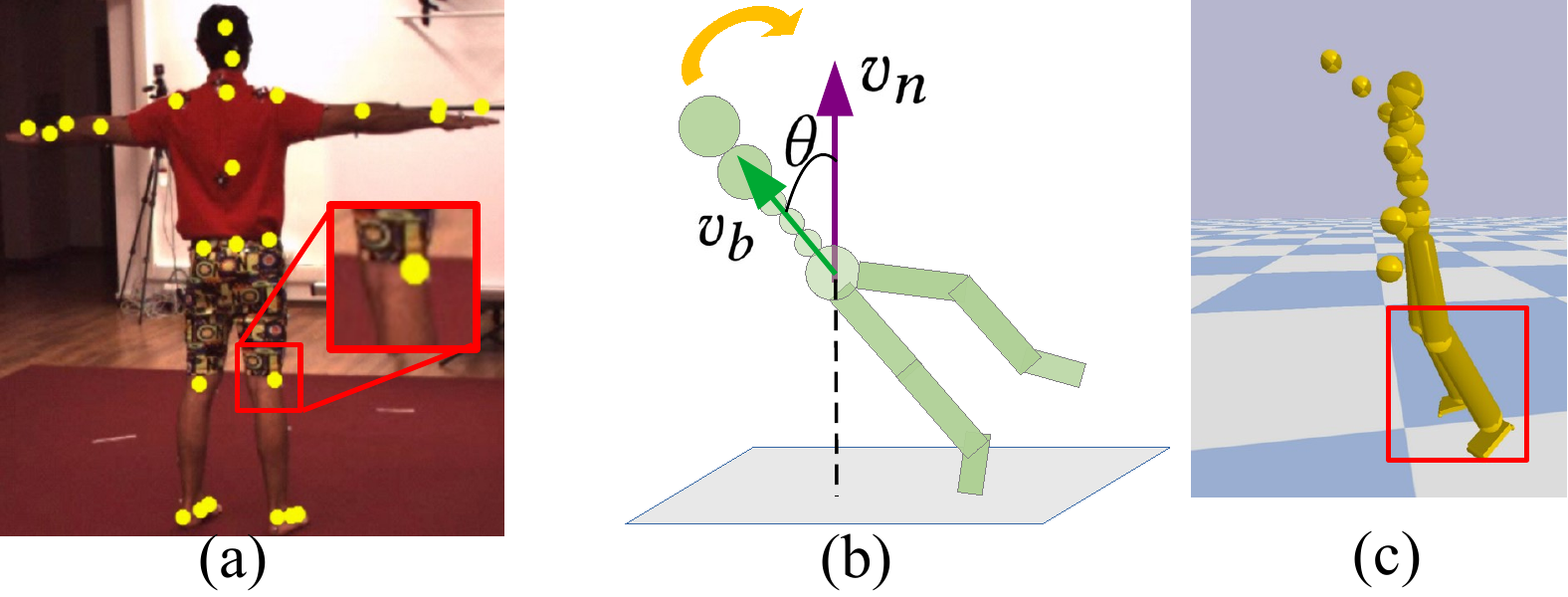} 
\caption{\label{fig:refCorrection}(a) An exemplary frame from the Human 3.6M dataset with the ground truth reprojections of the 3D joint keypoints. 
The magnified view in the red rectangle shows the reprojected keypoint that deviates from the rotation centre (the middle of the knee). 
(b) Schematic visualisation of the reference motion correction. 
Readers are referred to Sec.~\ref{ssec:initial_pose_correction} for its  details. 
(c) Example of a visually unnatural standing (stationary) pose caused by physically implausible knee bending.} 
\end{figure} 
the forefoot and heel joint contact labels are computed based on the assumption that a joint in contact is not sliding, \textit{i.e.,} the velocity is lower than 
$5$ cm/sec. 
In addition, we use a height criterion, \textit{i.e.,} the forefoot/heel, when in contact with the floor, has to be at a 3D height that is lower than a threshold $h_{\text{thres.}}$.
To determine this threshold for each sequence, we calculate the average heel $h^{heel}_{\text{avg}}$ and forefoot $h^{ffoot}_{\text{avg}}$ heights for each subject using the first ten frames (when both feet touch the ground). 
Thresholds are then computed as $h^{heel}_{\text{thres.}} = h^{heel}_{\text{avg}} + 5$cm for heels and $h^{ffoot}_{\text{thres.}} = h^{ffoot}_{\text{avg}} + 5$cm for the forefeet. 
This second criterion is needed since, otherwise, a foot in the air  that is kept static could also be labelled as being in contact. 

We also automatically label stationary and non-stationary poses on the same sequences. When standing and walking, the CoG of the human body typically lies close to the pelvis in 3D, which corresponds to the skeletal root position in both the Human3.6M and DeepCap datasets. 
Therefore, when the velocity of 3D root is lower than a threshold $\varphi_{v}$, we classify the pose as \textit{stationary}, and \textit{non-stationary} otherwise. 
In total, around $600k$ sets of contact and motion state labels for the human images are generated.
 
\subsection{Stage III: Physically Plausible Global 3D Pose Estimation}\label{sec:method_stage_3}
Stage III uses the results of stages I and II as inputs, \textit{i.e.,} $\mathbf{q}_{kin}^t$ and $\boldsymbol{b}_t$.
It transforms the kinematic motion estimate into a physically plausible global 3D pose sequence that corresponds to the images and adheres to anatomy and environmental constraints imposed by the laws of physics. 
To this end, we represent the human as a torque-controlled simulated character with a floating base and PD joint controllers  \cite{SalemPD}. The core is to solve an energy-based optimisation problem to find the vector of forces $\boldsymbol{\tau}$ and accelerations $\ddot{\mathbf{q}}$ of the character such that the equations of motion with constraints are fulfilled (Sec.~\ref{ssec:character_torque_control}). 
This optimisation is preceded by several preprocessing steps applied to each frame. 
First i), we correct $\mathbf{q}_{kin}^t$ if it is strongly implausible based on several easy-to-test criteria (Sec.~\ref{ssec:initial_pose_correction}). 
Second ii), we estimate the desired acceleration $\ddot{\mathbf{q}}_{des}\in\mathbb{R}^{m}$ necessary to reproduce $\mathbf{q}^{t}_{kin}$ based on the PD control rule (Secs.~\ref{ssec:set_PD_controller}).
Third iii), in input frames in which a foot is in contact with the floor (Sec.~\ref{ssec:foot_floor_collision_detection}), we estimate the ground reaction force (GRF) $\mathbf{\lambda}$ (Sec.~\ref{ssec:GRF_estimation}). 
Fourth iv), we solve the optimisation problem~\eqref{eq:qp_main} to estimate $\boldsymbol{\tau}$ and accelerations $\ddot{\mathbf{q}}$ where the equation of motion with the estimated GRF $\mathbf{\lambda}$ and the contact constraint to avoid foot-floor penetration (Sec.~\ref{ssec:character_torque_control}) are integrated as constraints. Note that the contact constraint is integrated only when the foot is in contact with the floor. Otherwise, only the equation of motion without GRF is introduced as a constraint in \eqref{eq:qp_main}. v) Lastly, the pose is updated using the finite-difference method (Eq.~\eqref{eq:findif}) with the estimated acceleration $ \ddot{\mathbf{q}}$. The steps ii) - v) are iterated $n=4$ times for each frame of video. 

As also observed by \cite{andrews2016real}, this two-step optimisation  iii) and iv) reduces direct actuation of the character's root as much as possible (which could otherwise lead to slightly unnatural locomotion), and explains the kinematically estimated root position and orientation by torques applied to other joints as much as possible when there is a foot-floor contact. Moreover, this two-step optimisation is computationally less expensive rather than estimating $\ddot { \mathbf{q} }$, $\boldsymbol{\tau}$ and $\lambda$ simultaneously \cite{zheng2013human}. Our algorithm thus finds a plausible balance between pose accuracy, physical accuracy, the naturalness of captured motion and real-time performance. 

\subsubsection{Pose  Correction}\label{ssec:initial_pose_correction} 
Due to the error accumulation in stage I (\textit{e.g.,} as a result of the deviation of 3D annotations from the joint rotation centres in the skeleton model, see Fig.~\ref{fig:refCorrection}-(a), as well as inaccuracies in the neural network predictions and skeleton fitting), the estimated 3D pose $\mathbf{q}_{kin}^t$ is often not physically plausible. 
Therefore, prior to torque-based optimisation, we pre-correct a pose  $\mathbf{q}_{kin}^t$ from stage I if it is 1) stationary and 2) unbalanced, \textit{i.e.,} the CoG projects outside the BoS. 
If both correction criteria are fulfilled, we compute the angle $\theta_t$ between the ground plane normal  $v_{n}$ and the vector $v_{b}$ that defines the direction of the  spine relative to the root in the local character's coordinate system (see Fig.~\ref{fig:refCorrection}-(b) for the schematic visualisation). 
We then correct the orientation of the virtual character towards a posture, for which CoG projects inside BoS. 
Correcting $\theta_t$ in one large step could lead to instabilities in physics-based pose optimisation. 
Instead, we reduce $\theta_t$ by a small rotation of the virtual character around its horizontal axis (\textit{i.e.,} the axis passing through the transverse plane of a human body) starting with the corrective angle $\xi_t = \frac{\theta_t}{10}$ for the first frame. 
Thereby, we accumulate the degree of correction in $\xi$ 
for the subsequent frames, \textit{i.e.,} $\xi_{t+1} = \xi_t + \frac{\theta_t}{10}$. 
Note that $\theta_t$ is decreasing for every frame and the correction step is performed for all subsequent frames until 1) the pose becomes non-stationary or 2) CoG projects inside BoS\footnote{either after the correction or already in $\mathbf{q}_{kin}^t$ provided by stage I}. 

However, simply correcting the spine orientation by the skeleton rotation around the horizontal axis can lead to implausible standing poses, since the knees can still be unnaturally bent for the obtained upright posture (see Fig.~\ref{fig:refCorrection}-(c) for an example). 
To account for that, we adjust the respective DoFs of the knees and hips such that the relative orientation between upper legs and spine, as well as upper and lower legs, are more straight. 
The hip and knee correction starts if both correction criteria are \textit{still} fulfilled and $\theta_t$ is \textit{already} very small. 
Similarly to the $\theta$ correction, we introduce  accumulator variables for every knee and every hip. 
The correction step for knees and hips is likewise performed until 1) the pose becomes non-stationary or 2) CoG projects inside BoS\footnotemark[1].

\subsubsection{Computing the Desired Accelerations}\label{ssec:set_PD_controller} 
To control the physics-based virtual character such that it reproduces the kinematic estimate $\mathbf{q}_{kin}^t$, we set the desired joint acceleration $\ddot{\mathbf{q}}_{des}$ following the PD controller rule: 
\begin{equation}\label{eq:PD_control}
    \ddot{\mathbf{q}}_{des} = \ddot{\mathbf{q}}^{t}_{kin} + k_{p}({\mathbf{q}}^{t}_{kin}-\mathbf{q}) + k_{d}(\dot{\mathbf{q}}^{t}_{kin}-\dot{\mathbf{q}}).
\end{equation}
The desired acceleration $\ddot{\mathbf{q}}_{des}$ is later used in the GRF estimation step (Sec.~\ref{ssec:GRF_estimation}) and the final pose optimisation (Sec.~\ref{ssec:character_torque_control}). Controlling the character motion on the basis of a PD controller in the system enables 
the character to exert torques $\boldsymbol{\tau}$ which reproduce the kinematic estimate $\mathbf{q}_{kin}^t$ while significantly mitigating undesired effects such as joint and base position jitter. %

\subsubsection{Foot-Floor Collision Detection}\label{ssec:foot_floor_collision_detection} 
To avoid foot-floor penetration in the final pose sequence and to mitigate contact position sliding, we integrate hard constraints in the physics-based pose optimisation to enforce zero velocity of forefoot and heel links in Sec.~\ref{ssec:character_torque_control}. 
However, these constraints can lead to unnatural motion in rare cases when the state prediction network may fail to estimate the correct foot contact states (\textit{e.g.,} when the foot suddenly stops in the air while walking). 
We thus update the contact state output of the state prediction network $\mathbf{b}_{t, j \in \{1, \hdots, 4\}}$, to yield $\mathbf{b}'_{t, j \in \{1, \hdots, 4\}}$ as follows:
\begin{equation} 
\small 
\mathbf{b}'_{t, j \in \{1, \hdots, 4\}} = 
  \begin{cases} 
    1, & \text{if ( $\mathbf{b}^{j} = 1$ and $h^{j} < \psi$ ) or} \\ 
      & \text{the $j$-th link collides with the floor plane}, \\ 
    0, & \text{otherwise.} 
  \end{cases} 
\end{equation} 
This means we consider a forefoot or heel link to be in contact only if its height $h^{j}$ is less than a threshold $\psi = 0.1$m above the calibrated ground plane. 

In addition, we employ the \textit{Pybullet}  \cite{coumans2016pybullet} physics engine to detect foot-floor collision for the left and right foot links. 
Note that combining the mesh collision information with the predictions from the state prediction network is necessary because 
1) the foot may not touch the floor plane in the simulation when the subject's foot is  actually in contact with the floor due to the inaccuracy of $\mathbf{q}^{t}_{kin}$, 
and 
2) the foot can penetrate into the mesh floor plane if the network misdetects the contact state when there is actually a foot contact in $\mathbf{I}_{t}$.

\subsubsection{Ground Reaction Force (GRF) Estimation}\label{ssec:GRF_estimation} 
We first compute the GRF $\mathbf{\lambda}$ -- when there is a contact between a foot and floor -- which best explains the motion of the root as coming from stage I. 
However, the target trajectory from stage I can be physically implausible, and we will thus eventually also require a \textit{residual} force directly applied on the root to explain the target trajectory; this force will be computed in the final optimisation.   
To compute the GRF, we solve the following minimisation problem: 
\begin{equation}\label{eq:optimal_f}
\begin{matrix}
    \underset{\mathbf{\lambda}}{\min } \lVert{\mathbf{M}}_{ 1 }{   \ddot{\mathbf{q}}_{des}}+\mathbf{c}_{1}(\mathbf{q},\dot{\mathbf{q}}) -{ \mathbf{J} }_{1}^{ T}{ \mathbf{G} \mathbf{\lambda}  }\rVert, \\ 
    \text{s.t.}\;~\mathbf{\lambda} \in F, 
    \end{matrix}
\end{equation}
where $\lVert \cdot \rVert$ denotes $\ell^2$-norm, and $\mathbf{M}_{1} \in \mathcal{R}^{6\times m}$ together with $\mathbf{J}^{T}_{1} \in \mathcal{R}^{6 \times6N_{c}}$ are the first six rows of $\mathbf{M}$ and $\mathbf{J}^{T}$ that correspond to the root joint, respectively. $\mathbf{c}_{1}\in \mathcal{R}^{6}$ denotes the first six elements of $\mathbf{c}$ (see Eq.\ref{eq:eom}), which also correspond to the root joint.
Since we do not consider sliding contact, the contact force $\mathbf{\lambda}$ has to satisfy friction cone constraints. 
Thus, we formulate a linearised friction cone constraint $F$.
That is,
\begin{equation} 
F^{j} = \Big\{\lambda^{j} \in \mathcal{R}^{3}|\lambda^{j}_{n} >0, \left|\lambda^{j}_{t} \right|\leq \bar{\mu}\lambda^{j}_{n},\left|\lambda^{j}_{b} \right|\leq \bar{\mu}\lambda^{j}_{n}  \Big\}, 
\end{equation} 
where $\lambda^{j}_{n}$ is a normal component, $\lambda^{j}_{t}$ and $\lambda^{j}_{b}$ are the tangential components of a contact force at the $j$-th contact position; $\mu$ is a friction coefficient which we set to $0.8$ and the friction coefficient of inner linear cone approximation reads $\bar{\mu} = \mu/\sqrt{2}$. 

The GRF $\mathbf{\lambda}$ is then integrated into the subsequent optimisation step \eqref{eq:qp_main} to estimate torques and accelerations of all joints in the body, including an additional residual direct root actuation component that is needed to explain the difference between the global 3D root trajectory of the kinematic estimate and the final physically correct result. 
The aim is to keep this direct root actuation as small as possible, which is best  achieved by a two-stage strategy that first estimates the GRF separately. 
Moreover, we observed this two-step optimisation enables faster computation than estimating $\lambda$, $\ddot{\mathbf{q}}$ and  $\boldsymbol{\tau}$ all at once. 
It is hence more suitable for our approach which aims at real-time operation. 

\subsubsection{Physics-Based Pose Optimisation} 
\label{ssec:character_torque_control}
In this step, we solve an optimisation problem to estimate $\boldsymbol{\tau}$ and $\ddot{\mathbf{q}}$ to track $\mathbf{q}^{t}_{kin}$ using the equation of motion \eqref{eq:eom} as a constraint. When  contact is detected (Sec.~\ref{ssec:foot_floor_collision_detection}), we integrate the estimated ground reaction force $\lambda$ (Sec.~\ref{ssec:GRF_estimation}) in the equation of motion. In addition, we introduce contact constraints to prevent foot-floor penetration and foot sliding when contacts are detected. 

Let $\dot{\mathbf{r}}_{j}$ be the velocity of the $j$-th contact link. 
Then, using the relationship between $\dot{\mathbf{r}}_{j}$ and $\dot{\mathbf{q}}$ \cite{featherstone2014rigid}, we can write:  
\begin{equation} \label{eq:lin_vel}
 \mathbf{J}_{j}\dot { \mathbf{q} }  =\dot { \mathbf{r}}_{j}. 
\end{equation} 
When the link is in contact with the floor, the velocity perpendicular to the floor has to be zero or positive to prevent penetration. 
Also, we allow the contact links to have a small tangential velocity $\sigma$ to prevent an immediate foot motion stop which creates visually unnatural motion. 
Our contact constraint inequalities read: 
\begin{equation} \label{eq:constraints} 
\begin{matrix} 
  0\leq \dot{ r  }^{n}_{j},  \quad   |\dot{ r }^{t}_{j}| \leq \sigma,\quad  \text{and}\quad |\dot{ r  }^{b}_{j}| \leq \sigma , \\ 
  \end{matrix} 
\end{equation} 
where $\dot{ r }^{n}_{j}$ is the normal component of  $\dot{\mathbf{r}}_{j}$, and $\dot{ r }^{t}_{j}$ along with  $\dot{r}^{b}_{j}$ are the tangential elements of  $\dot{\mathbf{r}}_{j}$. 

Using the desired acceleration $\ddot{\mathbf{q}}_{des}$ (Eq.~\eqref{eq:PD_control}), the equation of motion  \eqref{eq:eom}, optimal GRF $\mathbf{\lambda}$ estimated in \eqref{eq:optimal_f} and contact constraints  \eqref{eq:constraints}, we formulate the optimisation problem for finding the physics-based motion capture result as:
\begin{equation} \label{eq:qp_main} 
\begin{matrix} \underset{\ddot {\mathbf{ q} },\boldsymbol{\tau}}{\min} \lVert\ddot { \mathbf{q} } -{ \ddot { \mathbf{q} }  }_{des}\rVert+\ \lVert\boldsymbol{\tau}\rVert ,  \\ 
\text{s.t.}~\;\mathbf{M}\ddot { \mathbf{q} } -\boldsymbol{\tau} =\mathbf{J}^{T} \mathbf{G}\lambda - \mathbf{c}(\mathbf{q},\dot{\mathbf{q}}), \,\text{and} \\    
0\leq \dot{ r  }^{n}_{j},  \;  |\dot{ r }^{t}_{j}| \leq \sigma,\;  |\dot{ r }^{b}_{j}| \leq \sigma , \forall j.
\end{matrix} 
\end{equation} 
 
The first energy term forces the character to reproduce $\mathbf{q}^{t}_{kin}$. The second energy term is the regulariser that minimises $\boldsymbol{\tau}$ to prevent the overshooting, thus modelling natural human-like motion. 
 
After solving \eqref{eq:qp_main}, the character pose is updated by Eq.~\eqref{eq:findif}. We iterate the steps ii) - v) (see stage III in Fig.~\ref{fig:pipeline}) $n=4$ times, and stage III returns the $n$-th output from v) as the final character pose $\mathbf{q}^{t}_{phys}$. 
The final output of stage III is a sequence of joint angles and global root translations and rotations that explains the image observations, follows the purely kinematic reconstruction from stage I, yet is physically and anatomically plausible and temporally stable. 

\section{Results}\label{sec:results} 
\begin{table} 
 \caption{\label{tab:dataset} Names and duration of our six newly recorded outdoor sequences captured using SONY DSC-RX0 at $25$ fps. 
}
 \scalebox{1.00}{
 \begin{tabular}{ c|c|c  }\hline
	 	Sequence ID	   &   Sequence Name  & Duration [sec]  \\   \hline
     1 & \textit{building 1}        &    132  \\  
     2 & \textit{building 2}         &    90  \\  
     3 & \textit{forest}            &    105  \\   
     4 & \textit{backyard}          &    60   \\ 
     5 & \textit{balance beam 1}   &    21   \\ 
     6 & \textit{balance beam 2}    &    12   \\  \hline
\end{tabular}
}
 
\end{table}
We first provide implementation details of \physcap{} (Sec.~\ref{ssec:implementation}) and then demonstrate its qualitative state-of-the-art results  (Sec.~\ref{ssec:qualitative_results}). 
We next evaluate \physcap{}'s performance quantitatively (Sec.~\ref{ssec:quantitative_results}) and conduct a user study to assess the visual physical plausibility of the results  (Sec.~\ref{ssec:user_study}). 

We test \physcap{} on widely-used benchmarks  \cite{ionescu2013human3,mono-3dhp2017,deepcap} 
as well as on  \textit{backflip} and \textit{jump} sequences provided  by~\cite{Peng2018}. 
We also collect a new dataset with various challenging motions. 
It features six sequences in general scenes performed by two subjects\footnote{the variety of motions per subject is high; there are only two subjects in the new dataset due to COVID-19 related recording restrictions} recorded at $25$ fps. 
For the recording, we used SONY DSC-RX0, see Table \ref{tab:dataset} for more details on the sequences. 
\subsection{Implementation}\label{ssec:implementation}
Our method runs in real time ($25$~fps on average) on a PC with a Ryzen7 2700 8-Core Processor, 32 GB RAM and GeForce RTX 2070 graphics card. 
In stage I, we proceed from a freely available demo version of VNect \cite{VNect_SIGGRAPH2017}. 
\begin{figure}[t!]
\centering
\includegraphics[width=\linewidth]{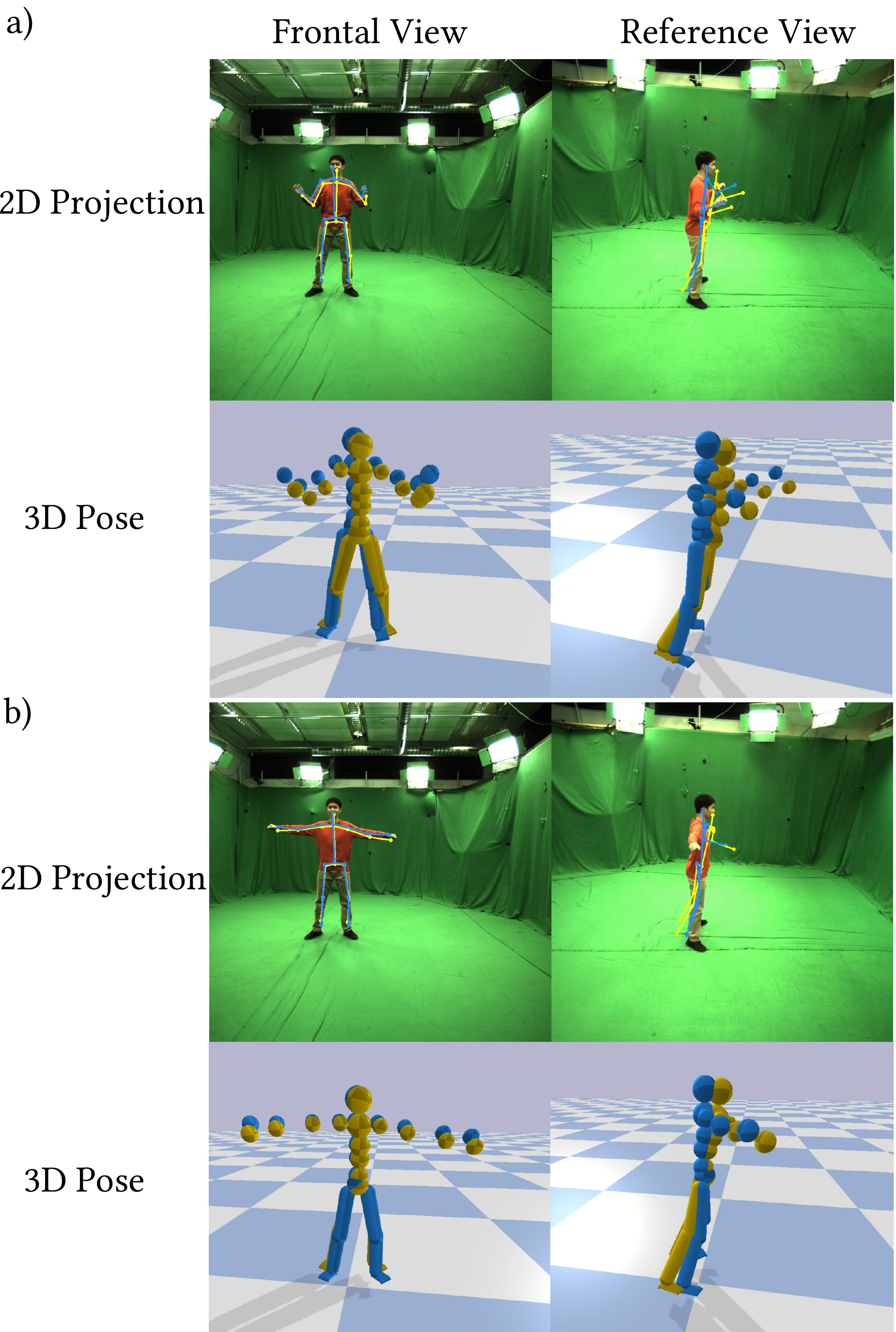}
\caption{ \label{fig:leaning} 
Two examples of reprojected 3D keypoints obtained by our approach (light blue colour) and Vnect \cite{VNect_SIGGRAPH2017} (yellow colour) together with 
the corresponding 3D visualisations from different view angles. 
\physcap{} produces much more natural and physically plausible postures whereas Vnect suffers from unnatural body leaning (see also the supplementary video). 
}
\end{figure}
Stages II and III are implemented in \textit{python}. 
In stage II, the network is implemented with \textit{PyTorch} \cite{NEURIPS2019_9015}. 
In stage III, we use the \textit{Rigid Body Dynamics Library} \cite{Felis2016} to compute dynamic quantities.
We employ the \textit{Pybullet} \cite{coumans2016pybullet} as a physics engine for the character motion visualisation and collision detection. 
In this paper, we set the proportional gain value $kp$ and derivative gain value $kd$ for all joints to $300$ and $20$, respectively. 
For the root angular acceleration, $kp$ and $kd$ are set to $340$ and $30$, respectively. 
$kp$ and $kd$ of the root linear acceleration are set to $1000$ and $80$, respectively. 
These settings are used in all experiments. 

\subsection{Qualitative Evaluation}\label{ssec:qualitative_results}

\begin{figure}[t!]
\centering
\includegraphics[width=\linewidth]{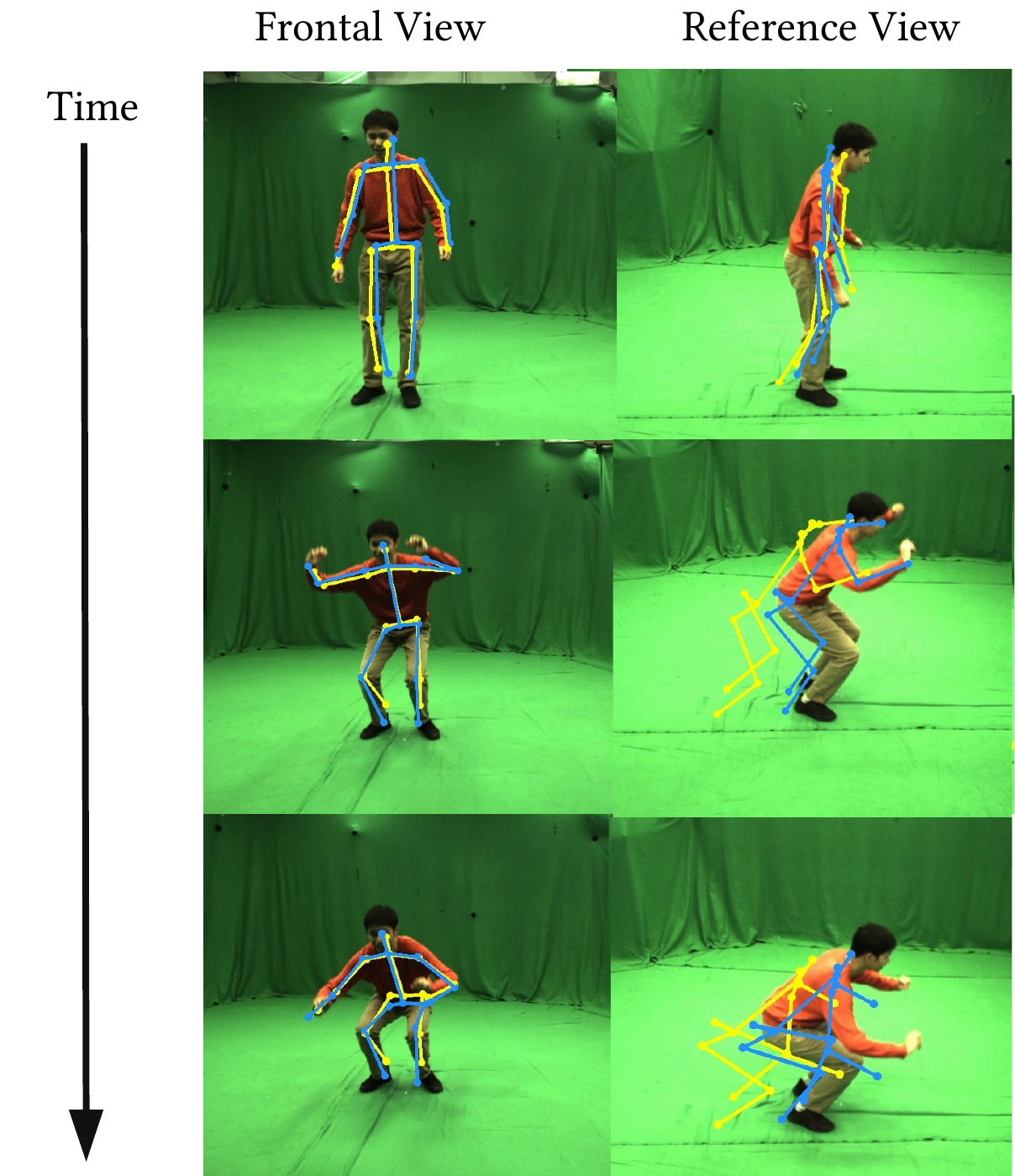} 
\caption{ \label{fig:sliding} Reprojected 3D keypoints onto two different images with different view angles for squatting. 
Frontal view images are used as inputs and images of the reference view are used only for quantitative evaluation. 
Our results are drawn in light blue, wheres the results by VNect \cite{VNect_SIGGRAPH2017} are provided in yellow. 
Our reprojections are more feasible, which is especially noticeable in the reference view. 
See also our supplementary video. 
} 
\end{figure}

\begin{figure}[t!]
\centering
\includegraphics[width=\linewidth]{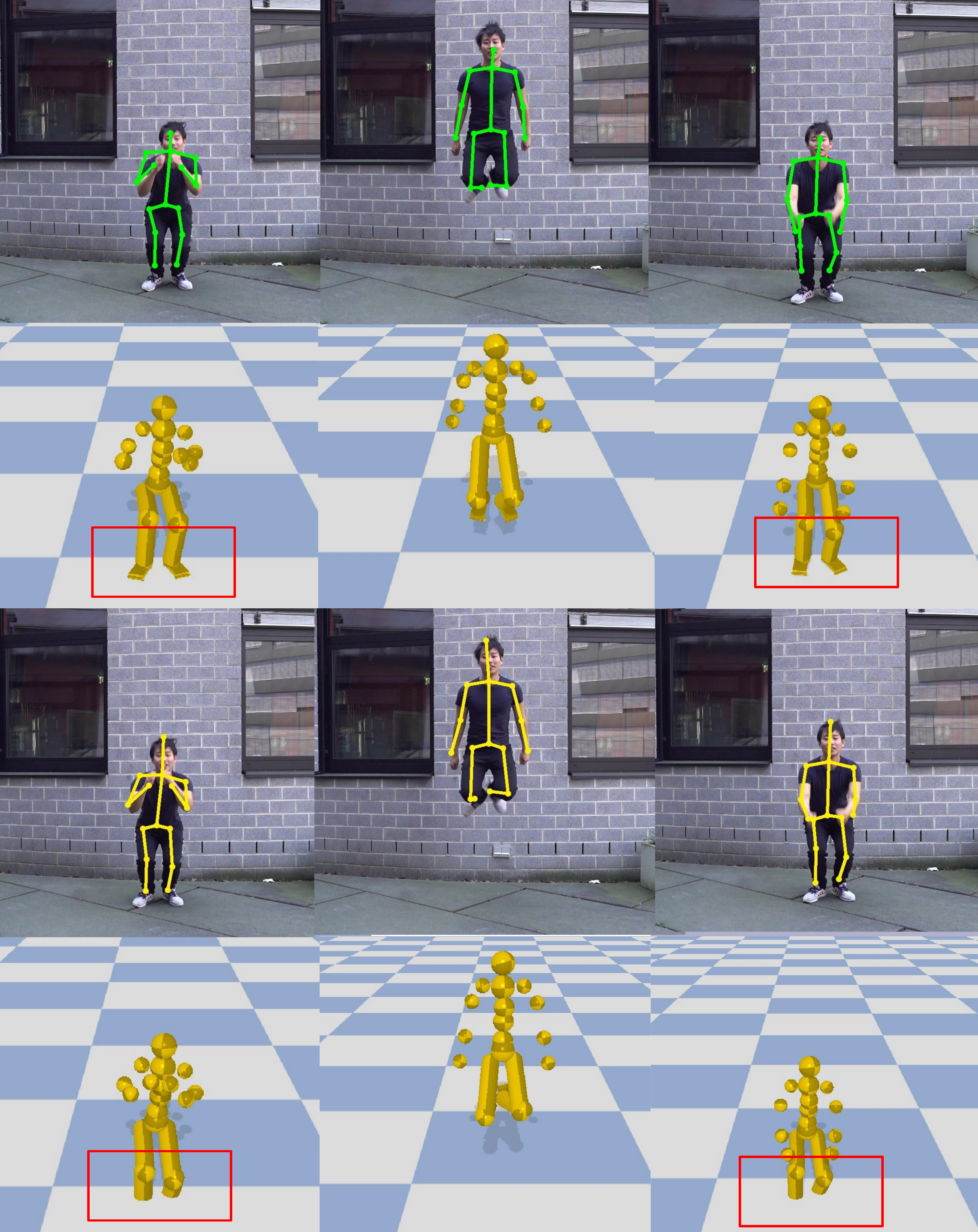}
\caption{ Several visualisations of the results by our approach and VNect \cite{VNect_SIGGRAPH2017}. 
The first and second rows show our estimated 3D poses after reprojection in the input image and its 3D view, respectively. 
Similarly, the third and fourth rows show the reprojected 3D pose and 3D view for VNect. 
Note that our motion capture shows no foot penetration into the floor plane whereas such artefact is apparent in the VNect  results. 
}
\label{fig:bigVisualization_penetration} 
\end{figure}

\begin{figure*}[t!]
\centering
\includegraphics[width= \linewidth]{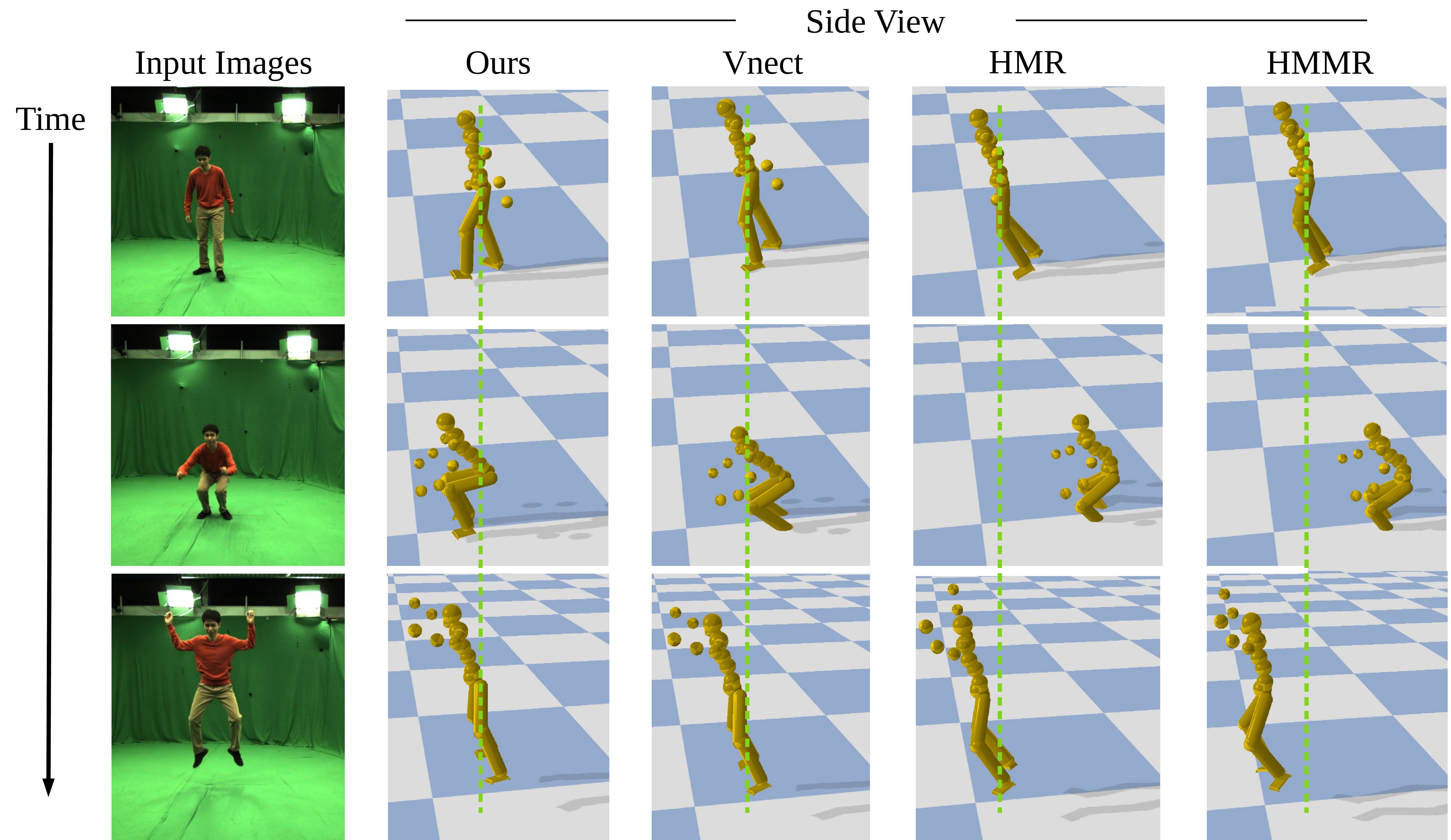}
\caption{\label{fig:side_related} Several side (non-input) view visualisations of the results by our approach, Vnect \cite{VNect_SIGGRAPH2017}, HMR \cite{hmrKanazawa17} and HMMR \cite{humanMotionKanazawa19} on DeepCap dataset. 
The green dashed lines indicate the expected root positions over time. 
It is apparent from the side view that our \physcap{} does not suffer from the unnatural body sliding along the depth direction, unlike other approaches. 
The global base positions for HMR and HMMR were computed by us using the root-relative predictions of these techniques, see Sec.~\ref{ssec:experimental_results} for more details. 
}
\end{figure*}

The supplementary video and result figures in this paper, in particular Figs.~\ref{fig:teaser} and  \ref{fig:bigVisualization_fancy} show that \physcap{} captures global 3D human poses in real time, even of fast and difficult motions, such as a backflip and a jump, which are of significantly improved quality compared to previous monocular methods. 
In particular, captured motions are much more temporally stable, and adhere to laws of physics with respect to the naturalness of body postures and fulfilment of environmental constraints, see Figs.~ \ref{fig:leaning}--\ref{fig:bigVisualization_penetration} and \ref{fig:side_related} for the examples of more natural 3D reconstructions. 
These properties are essential for many applications in graphics, in particular for stable real-time character animation, which is feasible by directly applying our method's output (see Fig.~\ref{fig:teaser} and the supplementary video).

\subsection{Quantitative Evaluation}\label{ssec:quantitative_results} 

In the following, we first describe our evaluation methodology in Sec.~\ref{ssec:evaluation_methodology}. 
We evaluate \physcap{} and competing methods under a variety of criteria, \textit{i.e.,} 3D joint position, reprojected 2D joint positions, foot penetration into the floor plane and motion jitter. 
We compare our approach with current state-of-the-art monocular pose estimation methods,\textit{i.e.,} HMR \cite{hmrKanazawa17}, HMMR \cite{humanMotionKanazawa19} and Vnect \cite{VNect_SIGGRAPH2017} (here we use the so-called \textit{demo version} provided by the authors with further improved accuracy over the original paper due to improved training). 
For the comparison, we use the benchmark dataset Human3.6M  \cite{ionescu2013human3}, the DeepCap dataset \cite{deepcap} and MPI-INF-3DHP \cite{mono-3dhp2017}.
From the Human3.6M dataset, we use the subset of actions that does not have occluding objects in the frame, \textit{i.e.,} \textit{directions, discussions, eating, greeting, posing, purchases, taking photos, waiting, walking, walking dog} and \textit{walking together}. 
From the DeepCap dataset, we use the subject 2 for this comparison. 

\subsubsection{Evaluation Methodology}\label{ssec:evaluation_methodology} 

The established evaluation methodology in monocular 3D human pose estimation and capture consists of testing a method on multiple sequences and reporting the accuracy of 3D joint positions as well as the accuracy of the reprojection into the input views. 
The accuracy in 3D is evaluated by \textit{mean per joint position error} (MPJPE) in mm, \textit{percentage of correct keypoints} (PCK) and the \textit{area under the receiver operating characteristic (ROC) curve} abbreviated as AUC. 
The reprojection or mean pixel error $e_{2D}^{\text{input}}$ is obtained by projecting the estimated 3D joints onto the input images and taking the average per frame distance to the ground truth 2D joint positions. 
We report $e_{2D}^{\text{input}}$ and its standard deviation denoted by $\sigma_{2D}^{\text{input}}$ with the images of size $1024 \times 1024$ pixels.
As explained earlier, these metrics only evaluate limited aspects of captured 3D poses and do not account for essential aspects of temporal stability, smoothness and physical plausibility in reconstructions such as
jitter, foot sliding, foot-floor penetration and unnaturally balanced postures. 
As we show in the supplemental video, top-performing methods on MPJPE and 3D PCK can fare poorly with respect to these criteria. 
Moreover, MPJPE and PCK are often reported after rescaling of the result in 3D or Procrustes alignment, which further makes these metrics agnostic to the aforementioned artefacts.
Thus, we introduce four additional metrics which allow to evaluate the physical plausibility of the results, \textit{i.e.,} \textit{reprojection error to unseen views} $e_{2D}^\text{side}$, \textit{motion jitter error} $e_{smooth}$ and two floor penetration errors -- \textit{Mean Penetration Error (MPE)} and \textit{Percentage of Non-Penetration (PNP)}. 

When choosing a reference side view for $e_{2D}^\text{side}$, we make sure that the viewing angle between the input and side views has to be sufficiently large, \textit{i.e.,} more than ${\sim}\frac{\pi}{15}$. 
Otherwise, if a side view is close to the input view, such effects as unnatural leaning forward can still remain undetected by $e_{2D}^\text{side}$ in some cases. 
After reprojection of a 3D structure to an image plane of a side view, all further steps for calculating $e_{2D}^\text{side}$ are similar to the steps for the standard reprojection error.
We also report $\sigma_{2D}^\text{side}$, \textit{i.e.,} the standard deviation of $e_{2D}^\text{side}$. 
To quantitatively compare the motion jitter, we report the deviation of the temporal  consistency from the ground truth 3D pose. 
Our smoothness error $e_{smooth}$ is computed as follows: 
\begin{equation} \label{eq:jitter} 
\begin{matrix} 
Jit_{X} =  \rVert \mathbf{p}^{s,t}_{X} - \mathbf{p}^{s,t-1}_{X}\lVert,\\ 
Jit_{GT} =  \rVert \mathbf{p}^{s,t}_{GT} - \mathbf{p}^{s,t-1}_{GT}\lVert,\\ 
e_{smooth} = \frac{1}{Tm}\sum_{t=1}^{T}\sum_{s=1}^{m}|Jit_{GT}-Jit_{X}|, 
\end{matrix} 
\end{equation} 
where $\mathbf{p}^{s,t}$ represents the 3D position of joint $s$ in the time frame $t$. 
$T$ and $m$ denote the total numbers of frames in the video sequence and target 3D joints, respectively. 
The subscripts $X$ and $GT$ stand for the predicted output and ground truth, respectively. 
A lower $e_{smooth}$ indicates lower motion jitter in the predicted motion sequence. 
\begin{table*}[t!]
\center
\caption{\label{tab:3Derror} 3D error comparison on benchmark datasets with VNect \cite{VNect_SIGGRAPH2017}, HMR \cite{hmrKanazawa17} and HMMR \cite{humanMotionKanazawa19}. We report the MPJPE in mm, PCK at 150 mm and AUC. Higher AUC and PCK are better, and lower MPJPE is better. Note that the global root positions for HMR and HMMR were estimated by solving optimisation with a 2D projection loss using the 2D and 3D keypoints obtained from the methods. Our method is on par with and often close to the best-performing approaches on all datasets. It consistently produces the best global root trajectory. As indicated in the text, these widely-used metrics in the pose estimation literature only paint an incomplete picture. For more details, please refer Sec.~\ref{ssec:quantitative_results}. 
}
 \scalebox{0.85}{
 \begin{tabular}{ c c|c c c|c c c|c c c }\hline
 
 				 &    &  \multicolumn{3}{c|}{DeepCap}     &  \multicolumn{3}{c| }{Human 3.6M}  &  \multicolumn{3}{c }{MPI-INF-3DHP}  \\   
				  &   &  MPJPE [mm] $\downarrow$ &  PCK[\%] $\uparrow$  &  AUC[\%]$\uparrow$  &  MPJPE [mm] $\downarrow$ &  PCK[\%] $\uparrow$ &  AUC[\%]$\uparrow$ &  MPJPE [mm] $\downarrow$ &  PCK[\%] $\uparrow$  &  AUC[\%]$\uparrow$ \\   \hline
 
 \multirow{4}{*}{Procrustes}    &  ours		    & 68.9     & \bf{95.0}  & 57.9    & 65.1   & 94.8    & 60.6 & 104.4 & 83.9 & 43.1 \\  
     &  Vnect   & \textbf{68.4}& 94.9	 &  \textbf{58.3} & 62.7   & 95.7     & 61.9  & 104.5 & 84.1 & 43.2  \\
     &   HMR    & 77.1    & 93.8  & 52.4     & \bf{54.3}   & 96.9     & \bf{66.6}  &\bf{87.8} & \bf{87.1} & \bf{50.9}  \\ 
     &  HMMR    & 75.5		& 93.8	  & 53.1     & 55.0   & \bf{96.6}    & 66.2  &106.9&79.5&44.8  \\ \hline
  \multirow{4}{*}{no Procrustes}    &   ours		    & 113.0     & 75.4   & 39.3     & 97.4   & 82.3     & 46.4 & 122.9&72.1 &35.0  \\ 
    &   Vnect    & 102.4		& 80.2	 & \textbf{42.4} & 89.6   & 85.1     & 49.0& \bf{120.2} & \bf{74.0} & \bf{36.1} \\
    &   HMR	    & 113.4     & 75.1  & 39.0     & \bf{78.9}   & 88.2     & \bf{54.1}  &130.5&69.7&35.7 \\
     &  HMMR     & \textbf{101.4}	& \textbf{81.0}	  & 42.0    & 79.4   & \bf{88.4}     & 53.8  &174.8&60.4&30.8 \\ \hline
  \multirow{4}{*}{global root position}   & ours		    & \bf{110.5}     & \bf{80.4}   & \bf{37.0}     & \bf{182.6}   & \bf{54.7}     & \bf{26.8}   & \bf{257.0} & \bf{29.7} & \bf{15.3} \\ 
    &   Vnect       & 112.6		& 80.0	 & 36.8     & 185.1    &  54.1     &  26.5   &261.0& 28.8 &15.0 \\
    &   HMR	    & 251.4     & 19.5  & 8.4     & 204.2   & 45.8     & 22.1  & 505.0 & 28.6 & 13.5 \\
   &    HMMR     & 213.0		& 27.7	  & 11.3     & 231.1   & 41.6     & 19.4  &926.2 & 28.0 & 14.5 \\ \hline
\end{tabular}
}
 
\end{table*}
MPE and PNP measure the degree of non-physical foot penetration into the ground. 
MPE is the mean distance between the floor and 3D foot position, and it is computed only when the foot is in contact with the floor. 
We use the ground truth foot contact labels (Sec.~\ref{sec:method_stage_2}) to judge the presence of the actual foot contacts. 
The complementary PNP metric shows the ratio of frames where the feet are not below the floor plane over the entire sequence. 

\subsubsection{Quantitative Evaluation Results}\label{ssec:experimental_results} 
%


Table \ref{tab:3Derror} summarises MPJPE, PCK and AUC for root-relative joint positions with (first row) and without (second row) Procrustes alignment before the error computation for our and related methods. 
We also report the global root position accuracy in the third row. 
Since HMR and HMMR do not return global root positions as their outputs, we estimate the root translation in 3D by solving an optimisation with 2D projection energy term using the 2D and 3D keypoints obtained from these algorithms (similar to the solution in VNect). 
The 3D bone lengths of HMR and HMMR were rescaled so that they match the ground truth bone lengths. 

In terms of MPJPE, PCK and AUC, our method does not outperform the other approaches consistently but achieves an accuracy that is comparable and often close to the highest on Human3.6M, DeepCap and MPI-INF-3DHP. 
In the third row, we additionally evaluate the global 3D base position accuracy, which is critical for character animation from the captured data. Here, \physcap{} consistently outperforms the other methods on all the datasets. 

As noted earlier, the above metrics only paint an incomplete picture. 
\begin{table}
\center
\caption{\label{tab:projection} 2D projection error of a frontal view (input) and side view (non-input) on DeepCap dataset \cite{deepcap}. 
\physcap{} performs similarly to VNect on the frontal view, and significantly better on the side view. For  further details, see  Sec.~\ref{ssec:quantitative_results} and  Fig.~\ref{fig:sliding}  
} 
\scalebox{0.87}{
\begin{tabular}{ c|c c|c c }\hline
                    &  \multicolumn{2}{c|}{Front View}     &  \multicolumn{2}{c }{Side View}\\ 
				    &  $e_{2D}^{\text{input}}$ [pixel]  &  $\sigma_{2D}^{\text{input}}$  &  $e_{2D}^{\text{side}}$ [pixel]  &  $\sigma_{2D}^{\text{side}}$  \\   \hline
      Ours		    & 21.1     & 6.7  & \textbf{35.5}     & \textbf{16.8}  \\ 
      Vnect \cite{VNect_SIGGRAPH2017}       & \bf{14.3}		& \textbf{2.7} & 37.2     & 18.1 	 \\  \hline 
\end{tabular}
}
 
\end{table} 
Therefore, we also measure the 2D projection errors to the input and side views on the DeepCap dataset, since this dataset includes multiple synchronised views of dynamic scenes with a wide baseline.
Table \ref{tab:projection} summarises the mean pixel errors $e_{2D}^{\text{input}}$ and $e_{2D}^{\text{side}}$ together with their standard deviations.
In the frontal view, \textit{i.e.,} on $e_{2D}^{\text{input}}$, VNect has higher accuracy than \physcap{}.
However, this comes at the prize of frequently violating physics constraints (floor penetration) and producing unnaturally leaning and jittering 3D poses (see also the supplemental video). 
In contrast, since \physcap{} explicitly models physical pose plausibility, it excels VNect in the side view, which reveals VNect's implausibly leaning postures and root position instability in depth, also see Figs.~\ref{fig:leaning} and \ref{fig:sliding}. 

\begin{table} 
\center
\caption{\label{tab:jitter} Comparison of temporal smoothness on the DeepCap \cite{deepcap} and Human 3.6M datasets  \cite{ionescu2013human3}. 
\physcap{} significantly outperforms VNect and HMR, and fares comparably to HMMR in terms of this metric. 
For a detailed explanation, see  Sec.~\ref{ssec:quantitative_results}. 
}
 \scalebox{1.00}{
 \begin{tabular}{ c c|c|c|c|c }\hline
	 		  &   & Ours & Vnect & HMR & HMMR\\   \hline
   \multirow{2}{*}{DeepCap}	  &  $e_{smooth}$	 & \bf{6.3}     & 11.6	    & 11.7     & 8.1  \\ 
      & $\sigma_{smooth}$       & \bf{4.1}     & 8.6	    & 9.0      & 5.1  \\  \hline
   \multirow{2}{*}{Human 3.6M}  &  $e_{smooth}$	 &  7.2      & 11.2	    & 11.2     & \bf{6.8}  \\ 
      & $\sigma_{smooth}$       &  6.9      & 10.1	    & 12.7     & \bf{5.9}  \\  \hline
\end{tabular}
}
 
\end{table}
To assess motion smoothness, we report $e_{smooth}$ and its standard deviation $\sigma_{smooth}$ in Table \ref{tab:jitter}. 
Our approach outperforms Vnect and HMR by a big margin on both datasets. 
Our method is better than HMMR on DeepCap dataset and marginally worse on Human3.6M. HMMR is one of the current state-of-the-art algorithms that has an explicit temporal component in the architecture. 

\begin{table}
\center
\caption{\label{tab:penetration} Comparison of Mean Penetration Error (MPE) and Percentage of Non-Penetration (PNP) on DeepCap dataset \cite{deepcap}. \physcap{} significantly outperforms VNect on this metric, measuring an essential aspect of physical motion correctness. 
}
\scalebox{0.95}{
\begin{tabular}{ c|c c|c  }\hline
				    & MPE [mm] $\downarrow$ &$\sigma_{MPE}$ $\downarrow$ & PNP [\%] $\uparrow$ \\   \hline
      Ours		    & \textbf{28.0}     & \textbf{25.9} & \textbf{92.9}	        \\ 
      Vnect \cite{VNect_SIGGRAPH2017}       & 39.3		& 37.5   	& 45.6	 \\   \hline
\end{tabular}
}
 
\end{table}
\begin{figure}[t!]
\centering
\includegraphics[width=0.9\linewidth]{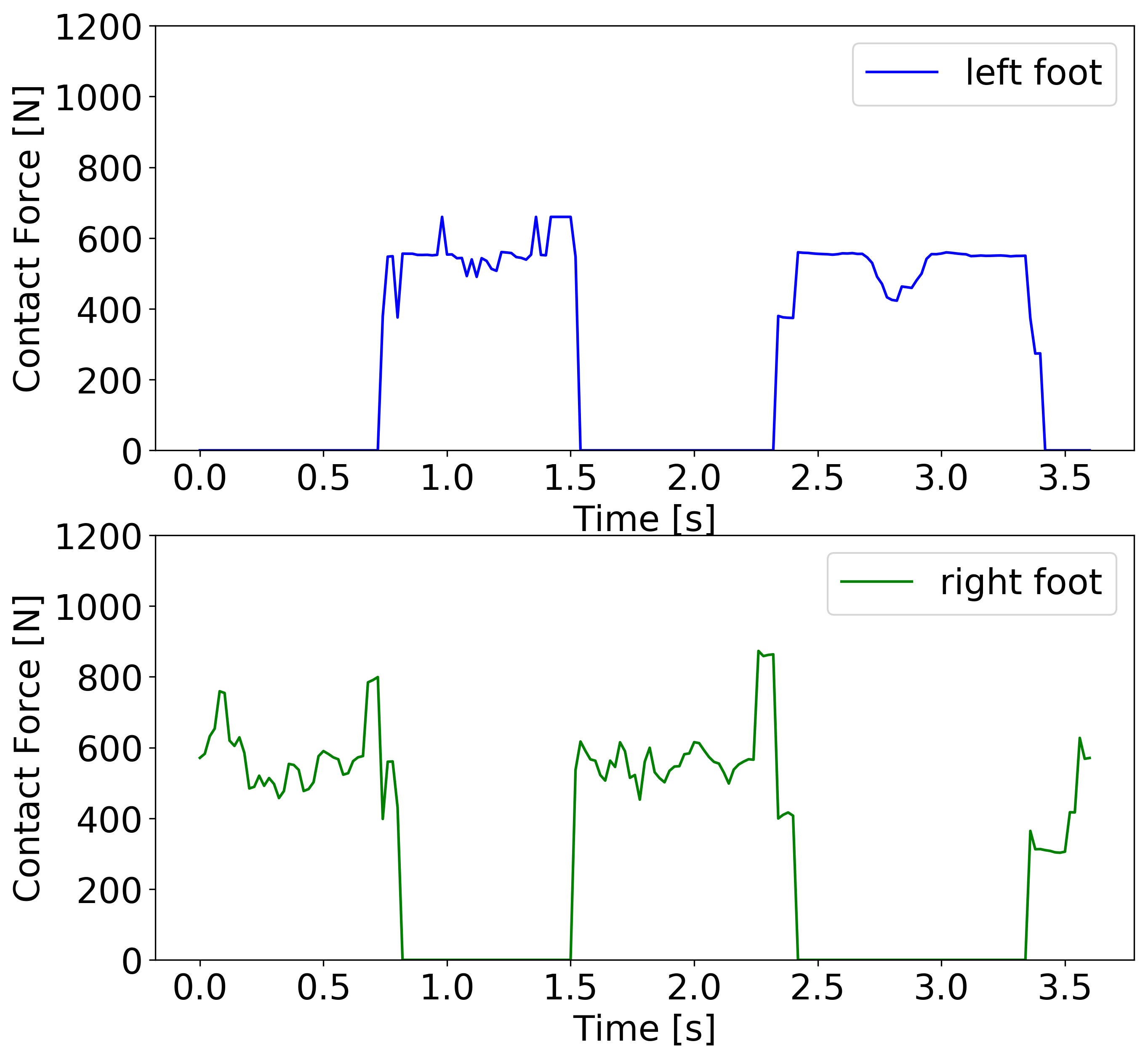}
\caption{\label{fig:contact_force} The estimated contact forces as the functions of time for the walking sequence. We observe that the contact forces remain in a reasonable range for walking motions \cite{shahabpoor2017measurement}.
}
\end{figure}

Table \ref{tab:penetration} summarises the MPE and PNP for Vnect and \physcap{} on DeepCap dataset. 
Our method shows significantly better results compared to VNect, \textit{i.e.,} about a $30\%$ lower MPE and a by $100\%$ better result in PNP, see  Fig.~\ref{fig:bigVisualization_penetration} for qualitative examples. 
Fig.~\ref{fig:contact_force} shows plots of contact forces as the functions of time calculated by our approach on the walking sequence from our newly recorded dataset (sequence 1). 
The estimated functions fall into a reasonable force range for walking motions \cite{shahabpoor2017measurement}.

\subsection{User Study}\label{ssec:user_study}
The notion of \textit{physical plausibility} can be understood and perceived subjectively from person to person. 
Therefore, in addition to the quantitative evaluation with existing and new metrics, we perform an online user study which allows to subjectively assess and compare the perceived degree of different effects in the reconstructions by a broad audience of people with different backgrounds in computer graphics and vision. 
In total, we prepared $34$ questions with videos, in which we always showed one or two reconstructions at a time (our result, a result by a competing method, or both at the same time). In total, $27$ respondents have participated. 
%
 
%
There were different types of questions. 
In $16$ questions (category I), the respondents were asked to decide which 3D reconstruction out of two looks more physically plausible to them (the first, the second or undecided). 
In $12$ questions (category II), the respondents were asked to rate how natural the 3D reconstructed motions are or evaluate the degree of an indicated effect (foot sliding, body leaning, \textit{etc.}) on a predefined scale. 
In five questions (category III), the respondents were also asked to decide which visualisation has a more pronounced indicated artefact. 
For two questions out of five, 2D projections onto the input 2D image sequence were shown, whereas the remaining questions in this category featured 3D  reconstructions.  
Finally (category IV), the participants were encouraged to list which artefacts in the reconstructions seem to be most apparent and most frequent.

\begin{figure*}
\centering
\includegraphics[width= 0.92\textwidth]{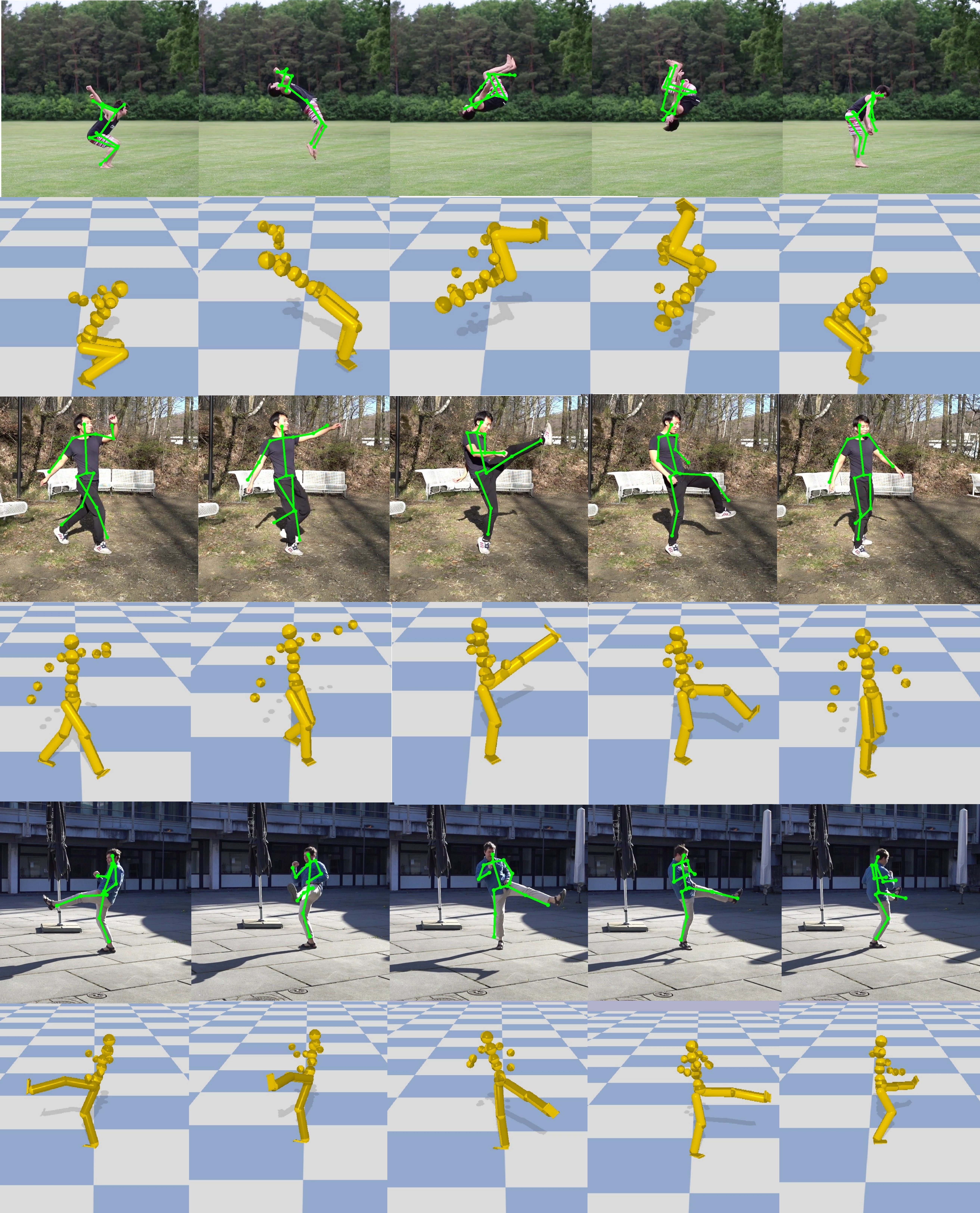}
\caption{ 
Representative 2D reprojections and the corresponding 3D poses of our \physcap{} approach. 
Note that, even with the challenging motions, our global poses in 3D have high quality and 2D reprojections to the input images are accurate as well. 
See our supplementary video for more results on these sequences. The \textit{backflip} video in the first row is taken from \cite{Peng2018}. Other sequences are from our own recordings. 
} 
\label{fig:bigVisualization_fancy} 
\end{figure*}

In category I, our reconstructions were preferred in $89.2\%$ of the cases, whereas a competing method was preferred in $1.6\%$ of the cases. 
Note that at the same time, the decision between the methods has not been made in $8.9\%$ cases. 
In category II, the respondents have also found the results of our approach to be significantly more physically plausible than the results of competing methods. 
The latter were also found to have consistently more jitter, foot sliding and unnatural body leaning. 
In category III, noteworthy is also that the participants have indicated a higher average perceived accuracy of our reprojections, \textit{i.e.,} $32.7\%$ voted that our results reproject better, whereas the choice felt on the competing methods in $22.6\%$ of the cases. 
Note that the smoothness and jitter in the results are also  reflected in the reprojections, and, thus, both influence how natural the reprojected skeletons look like. 
At the same time, a high uncertainty of $44.2\%$ indicates that the difference between the reprojections of \physcap{} and other methods is volatile. 
For the 3D motions in this category, $82.7\%$ voted that our results show fewer indicated artefacts compared to other approaches, whereas $13.5 \%$ of the respondents preferred the competing methods. The decision has not been made in $3.7 \%$ of the cases. 
In category IV, $59\%$ of the participants named jitter as the most frequent and apparent disturbing effect of the competing methods, followed by unnatural body leaning  ($22\%$), foot-floor penetration ($15\%$) and foot sliding ($15\%$). 

The user study confirms a high level of physical plausibility and naturalness of \physcap{} results. 
We see that also subjectively, a broad audience coherently finds our results of high visual quality, and the gap to the competing methods is substantial. 
This strengthens our belief about the suitability of \physcap{} for computer graphics and primarily virtual character animation in real time. 

\section{Discussion}\label{sec:discussion} 
Our physics-based monocular 3D human motion capture algorithm significantly reduces the common artefacts of other monocular 3D pose estimation methods such as motion jitter, penetration into the floor, foot sliding and unnatural body leaning. 
The experiments have shown that our state prediction network generalises  well across scenes with different backgrounds (see Fig.~\ref{fig:bigVisualization_fancy}). 
However, in the case of foot occlusion, our state prediction network can sometimes mispredict the foot contact states, resulting in the erroneous hard zero velocity constraint for feet. 
Additionally, our approach requires the calibrated floor plane to apply the foot contact constraint effectively; standard calibration techniques can be used for this. 
Swift motions can be challenging for stage I of our pipeline, which can cause inaccuracies in the estimates of the subsequent stages, as well as in the final estimate. 
In future, other monocular kinematic pose estimators than  \cite{VNect_SIGGRAPH2017} could be tested in stage I, in case they are trained to handle occlusions and very fast motions better. 
Moreover, note that -- although we use a single parameter set for \physcap{} in all our experiments (see Sec.~\ref{sec:results}) -- users can adjust the quality of the reconstructed motions by tuning the gain parameters of PD controller depending on the scenario. 
By increasing the derivative gain value, the reconstructed poses are  smoother, which, however, can cause motion delay compared to the input video, especially when the observed motions are very fast. 
By reducing the derivative gain value, our optimisation with a virtual  character can track image sequence with less motion delay, at the cost of less temporally coherent motion. We demonstrate this trade-off in the supplemental video. 

Further, while our method works in front of general backgrounds, we assume there is a ground plane in the scene, which is the case for most man-made environments, but not irregular outdoor terrains. 
Finally, our method currently only considers a subset of potential body-to-environment contacts in a physics-based way. As part of future work, we will investigate explicit modelling of self-collisions, as well as hand-scene interactions or contacts of legs and body in sitting and lying poses. 

\section{Conclusions}\label{sec:conclusions} 
We have presented \physcap{} -- the first physics-based approach for a global 3D human motion capture from a single RGB camera that runs in real time at $25$ fps. 
Thanks to the pose optimisation framework using PD joint control, the results of \physcap{} evince improved physical plausibility, temporal consistency and significantly fewer artefacts such as jitter, foot sliding, unnatural body leaning and foot-floor penetration, compared to other existing approaches (some of them include temporal constraints). 
We also introduced new error metrics to evaluate these improved properties which are not easily captured by metrics used in the established pose estimation benchmarks. 
Moreover, our user study further confirmed these improvements. 
In future work, our algorithm can be extended for various contact positions (not only the feet). 
%

%
%
%
%

\bibliographystyle{ACM-Reference-Format}
\bibliography{sample-bibliography}

\appendix 

\end{document}